\documentclass[11pt]{article}

\usepackage{zingpaper}
\usepackage{placeins}
\newcommand{\zingRtxGpuCount}{8}
\newcommand{\zingConcurrentStreams}{8}
\newcommand{\zingRtxResolution}{\ensuremath{832\times480}}
\newcommand{\zingPlaybackFPS}{24}
\newcommand{\zingLatentFramesPerBlock}{4}
\newcommand{\zingDecodedFramesPerBlock}{16}

\newcommand{\zingMeasuredFPS}{24.63}

\newcommand{\zingStreamMinuteCost}{0.009}

\papertitle{\texorpdfstring{\fontsize{20}{24}\selectfont Zing-0.5: Toward Playable Worlds\\with Real-Time Joint Action and Text Control}{Zing-0.5: Toward Playable Worlds with Real-Time Joint Action and Text Control}}
\paperauthors{Zing Team}
\paperaffiliations{SeedLeap.ai}
\papernote{Technical Report \quad September 15, 2026}
\paperrightlogo{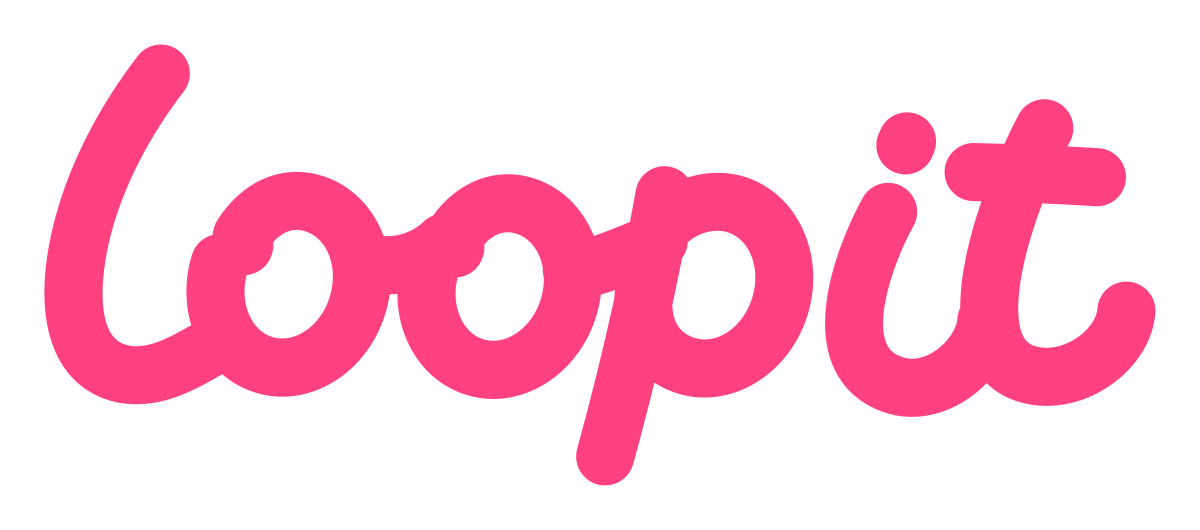}
\paperrighticon{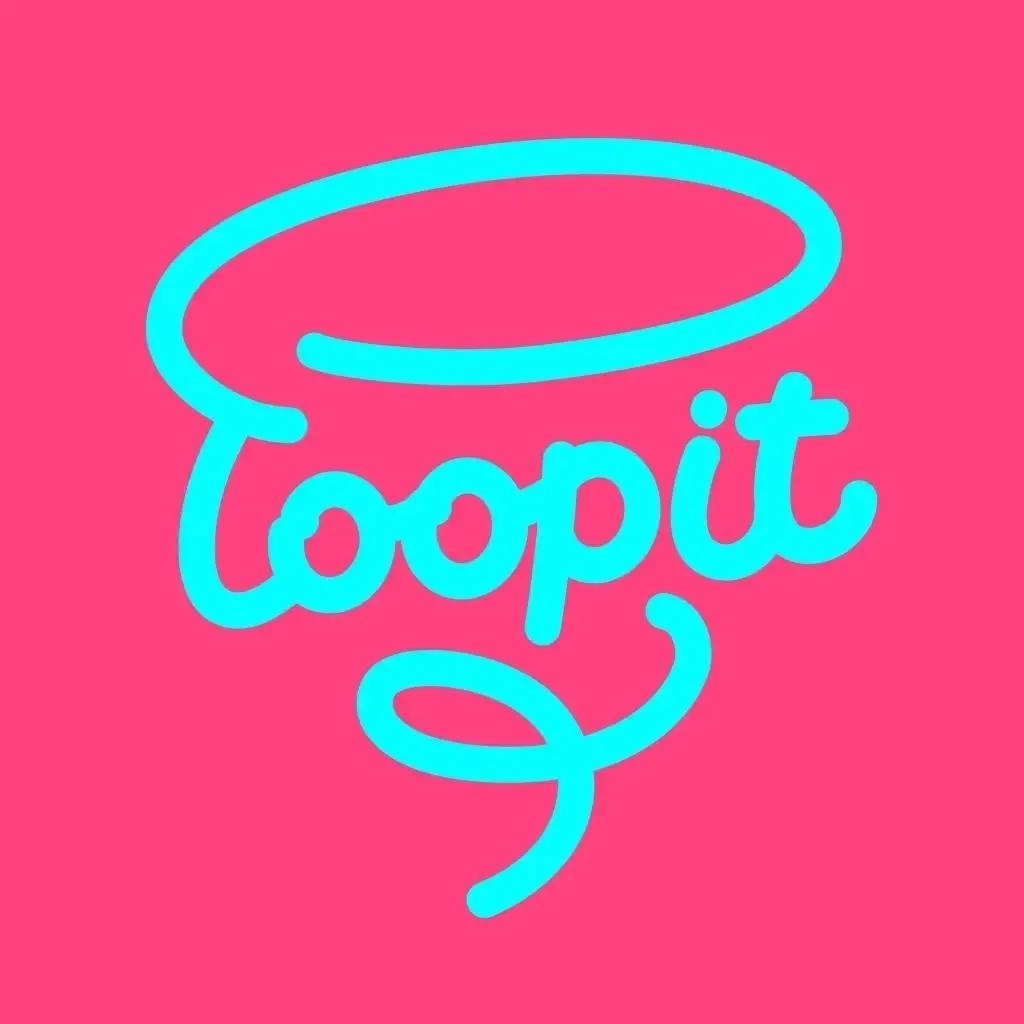}
\paperabstract{We introduce \mbox{Zing-0.5}, a 5B autoregressive world model designed for playability: users can explore generated worlds, influence unfolding events, and respond to the resulting feedback through joint keyboard and online text control. Our approach brings together three technical contributions: \textbf{(1) Unified action and text conditioning}, combining magnitude-aware keyboard inputs with temporally aligned text instructions and jointly annotated videos to learn navigation and event control within the same sequence; \textbf{(2) Event-scale supervision for incremental generation}, using a segment-level teacher trained on connected multi-prompt videos to supervise a block-level causal student through distribution-matching distillation; and \textbf{(3) Low-cost real-time interaction}, combining four-step generation with context-preserving streaming to support \zingRtxResolution{} inference at \zingPlaybackFPS{} FPS at an estimated server rental cost of approximately \$\zingStreamMinuteCost{} per stream-minute. \mbox{Zing-0.5} achieves an overall score of 81.0 and a consistency score of 88.5 across 158 WBench Navigation cases. A joint-control demonstration shows a text-directed event change during continued navigation without restarting generation. We release the model weights, inference code, and \href{https://github.com/seedleap/Zing-SGLang}{Zing-SGLang serving implementation} to support further work on playable generated worlds.}
\paperlinks{\paperlinkicon{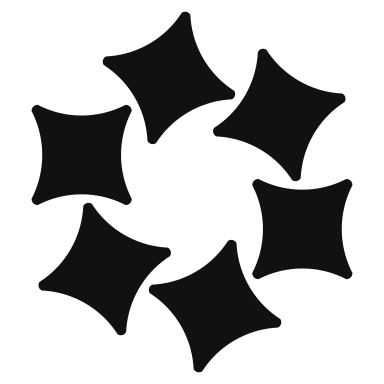}\textbf{Project:} \url{https://zing.loopit.me/}\par\paperlinkicon{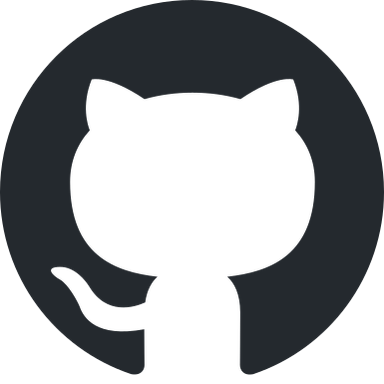}\textbf{Code:} \url{https://github.com/seedleap/zing-world-model}\par\paperlinkicon{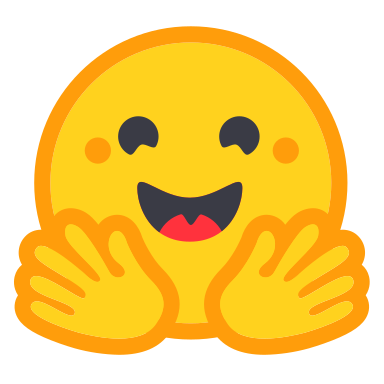}\textbf{Models:} \url{https://huggingface.co/seedleap/zing-0.5}}

\begin{document}

\makezingtitle

\begingroup
\setlength{\parskip}{0pt}
\newcommand{\teasertile}[1]{\includegraphics[width=\dimexpr(\linewidth-2.4mm)/4\relax]{figures/teaser/#1.jpg}}
\noindent\begin{minipage}{\linewidth}
\begingroup
\offinterlineskip
\hbox{%
  \teasertile{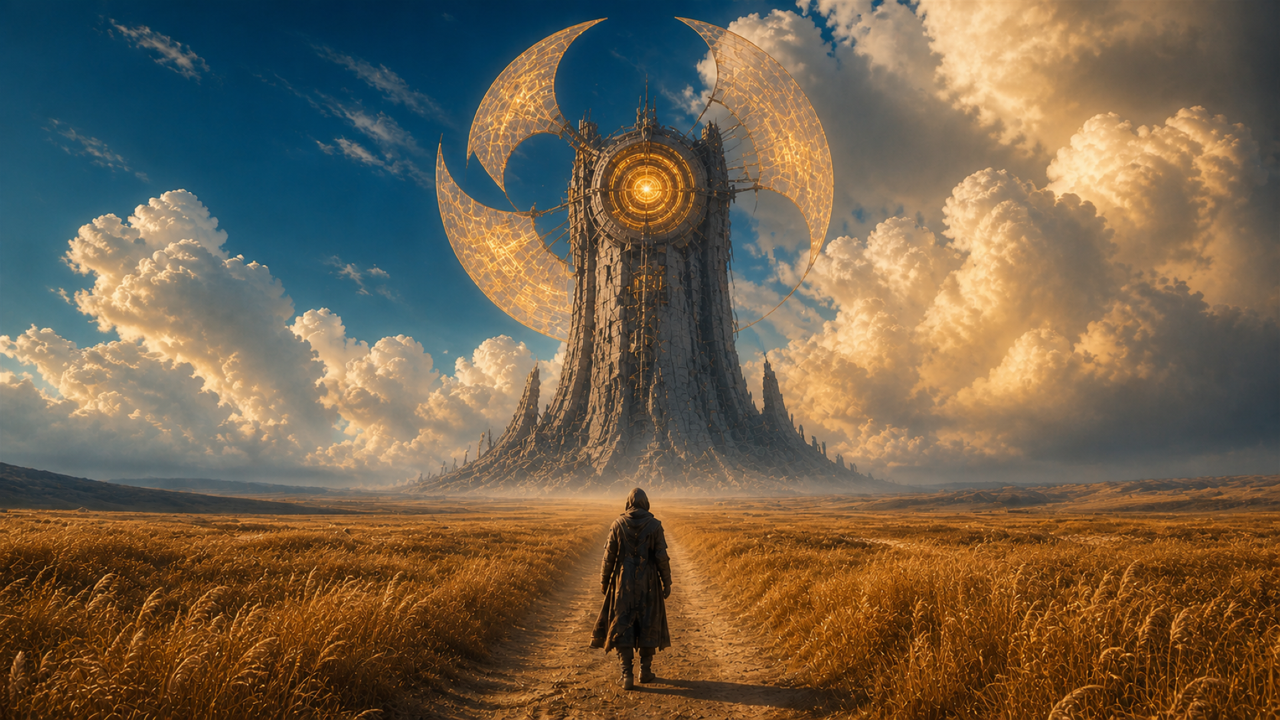}\hspace{0.8mm}%
  \teasertile{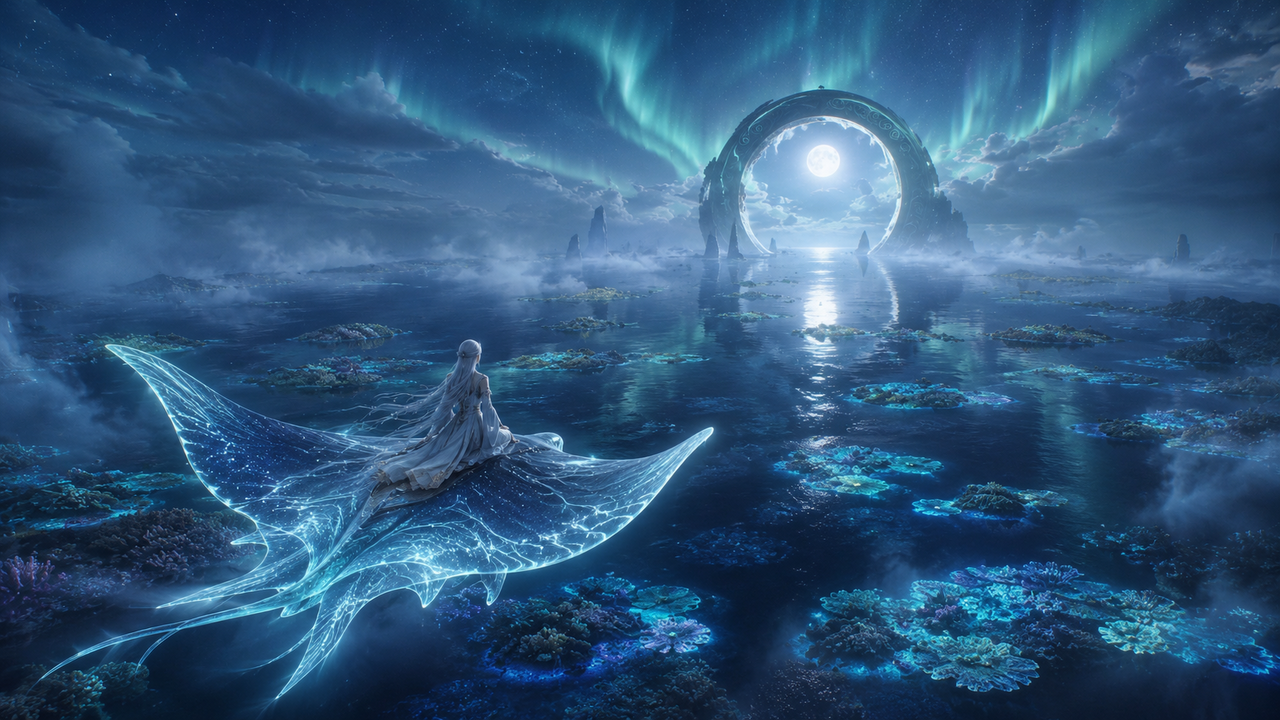}\hspace{0.8mm}%
  \teasertile{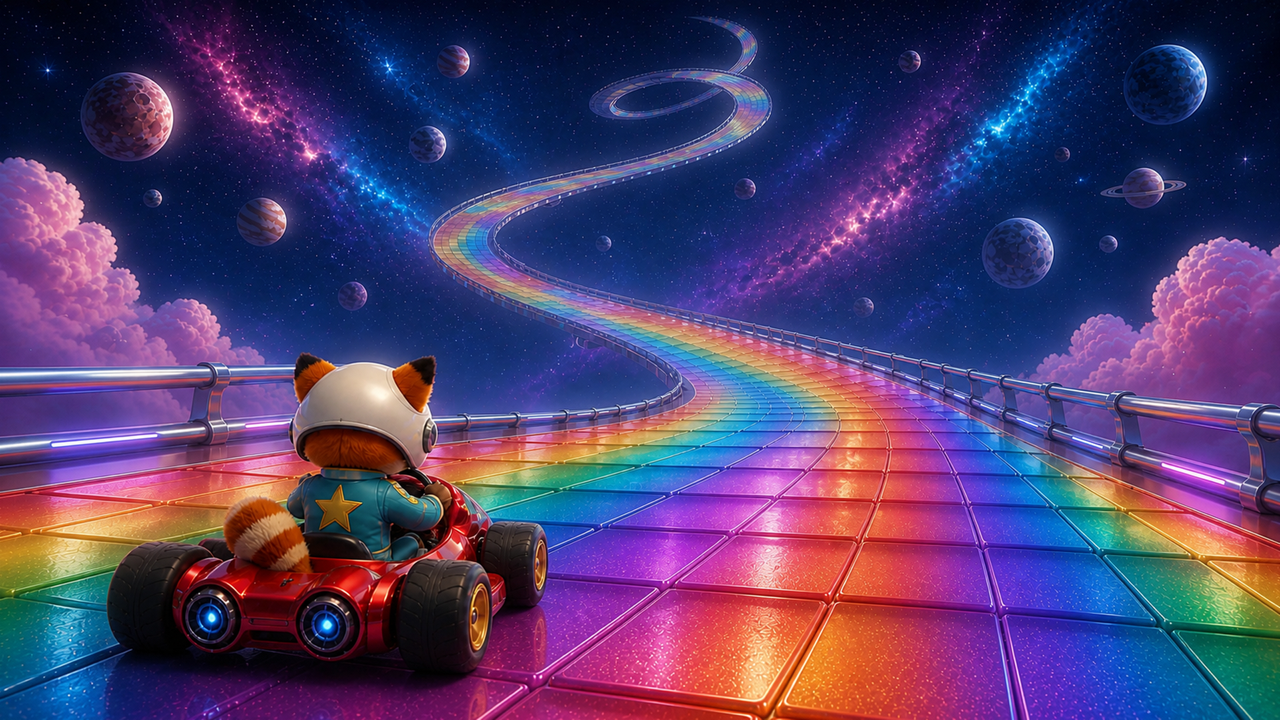}\hspace{0.8mm}%
  \teasertile{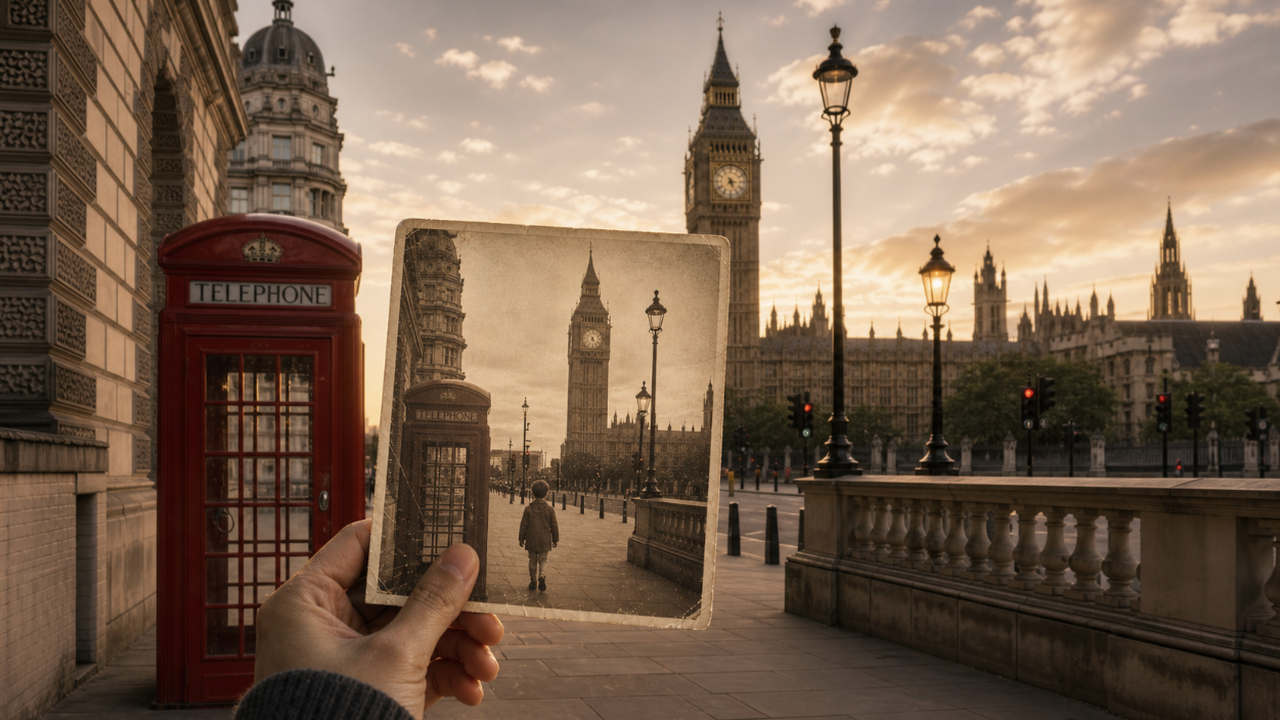}%
}
\vskip0.8mm
\hbox{%
  \teasertile{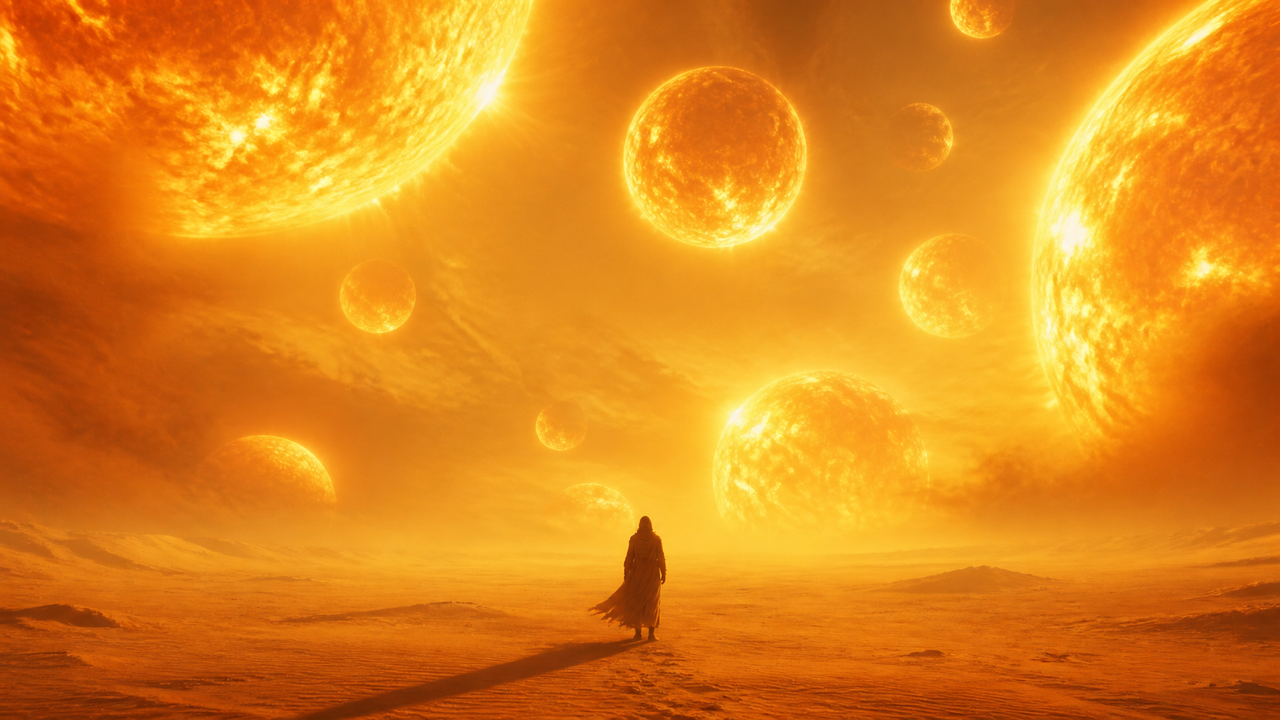}\hspace{0.8mm}%
  \teasertile{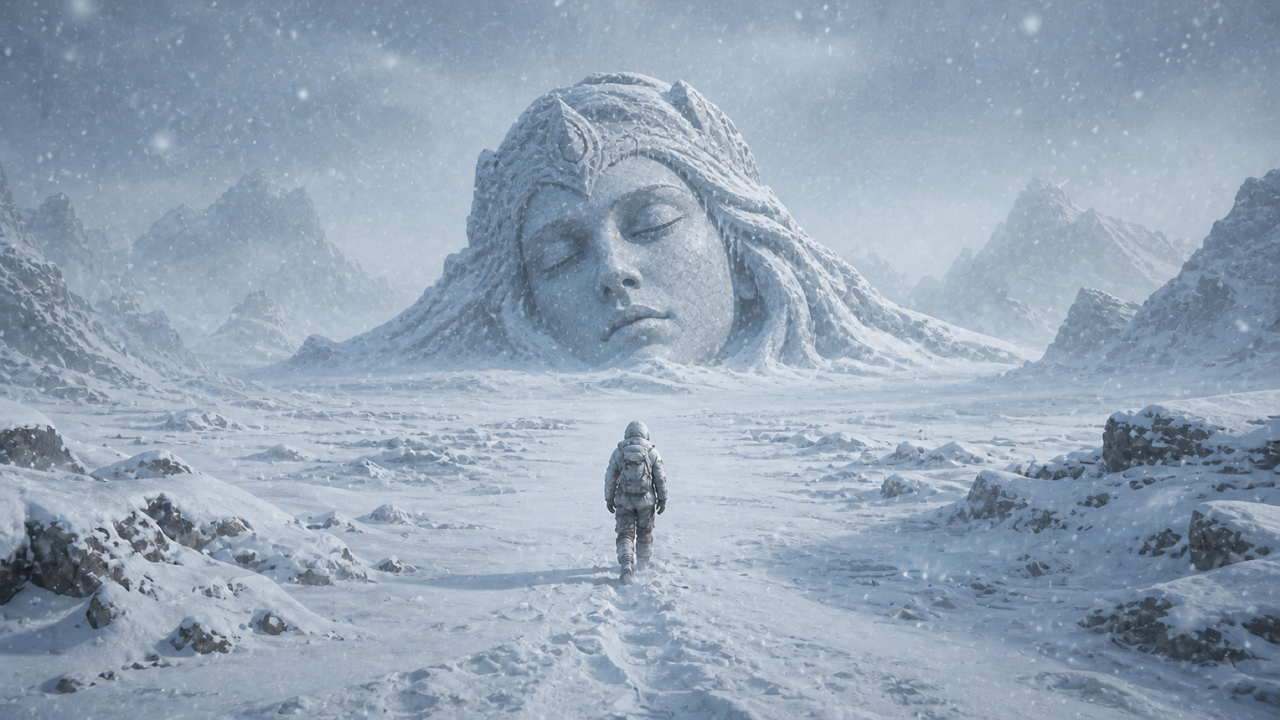}\hspace{0.8mm}%
  \teasertile{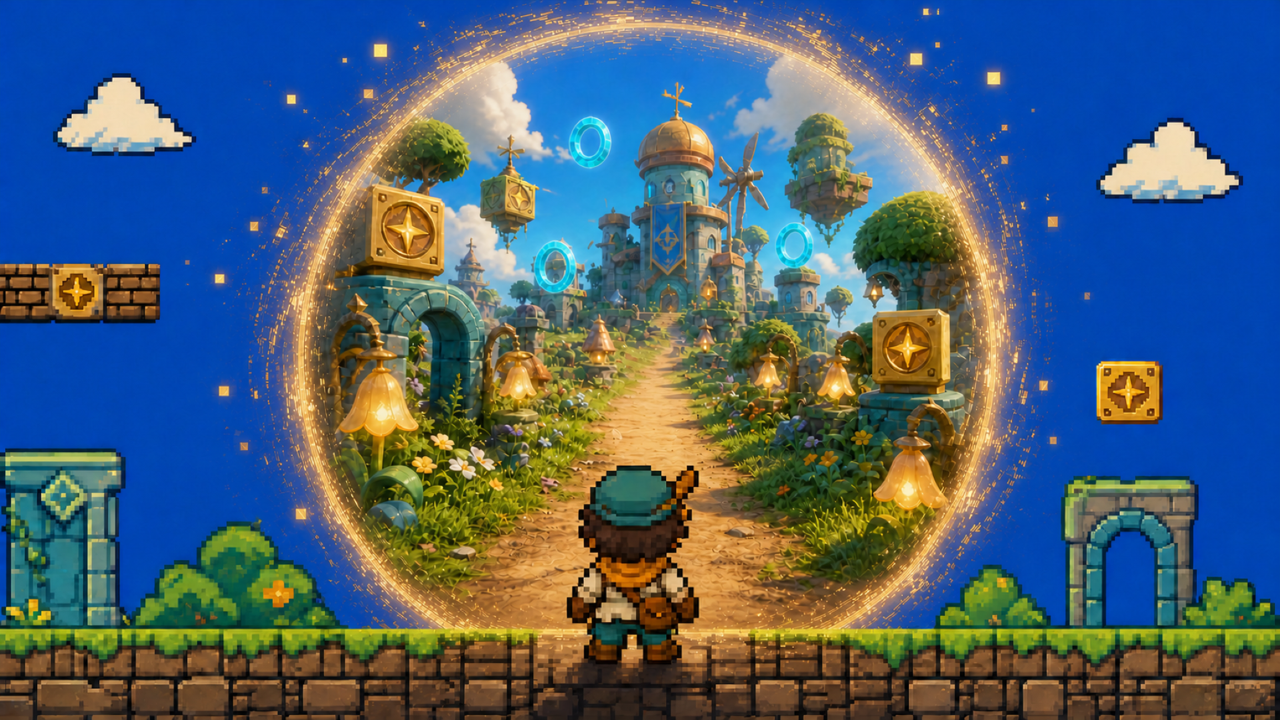}\hspace{0.8mm}%
  \teasertile{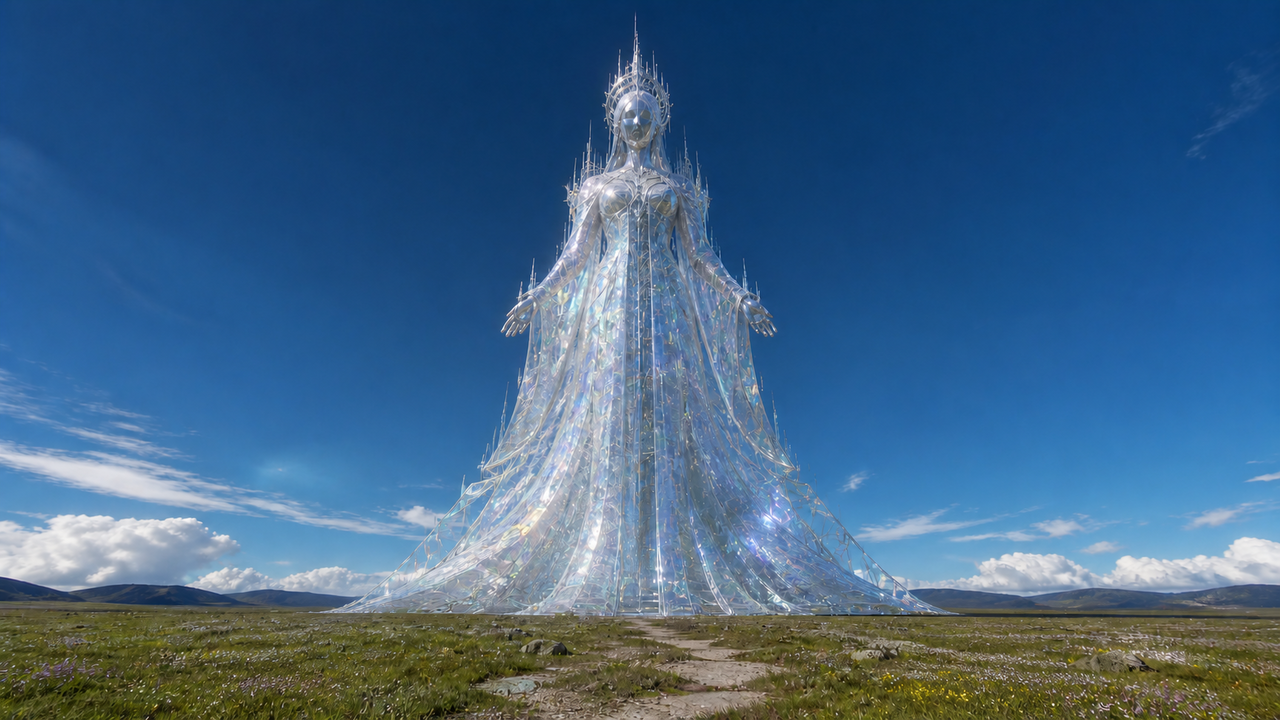}%
}
\vskip0.8mm
\hbox{%
  \teasertile{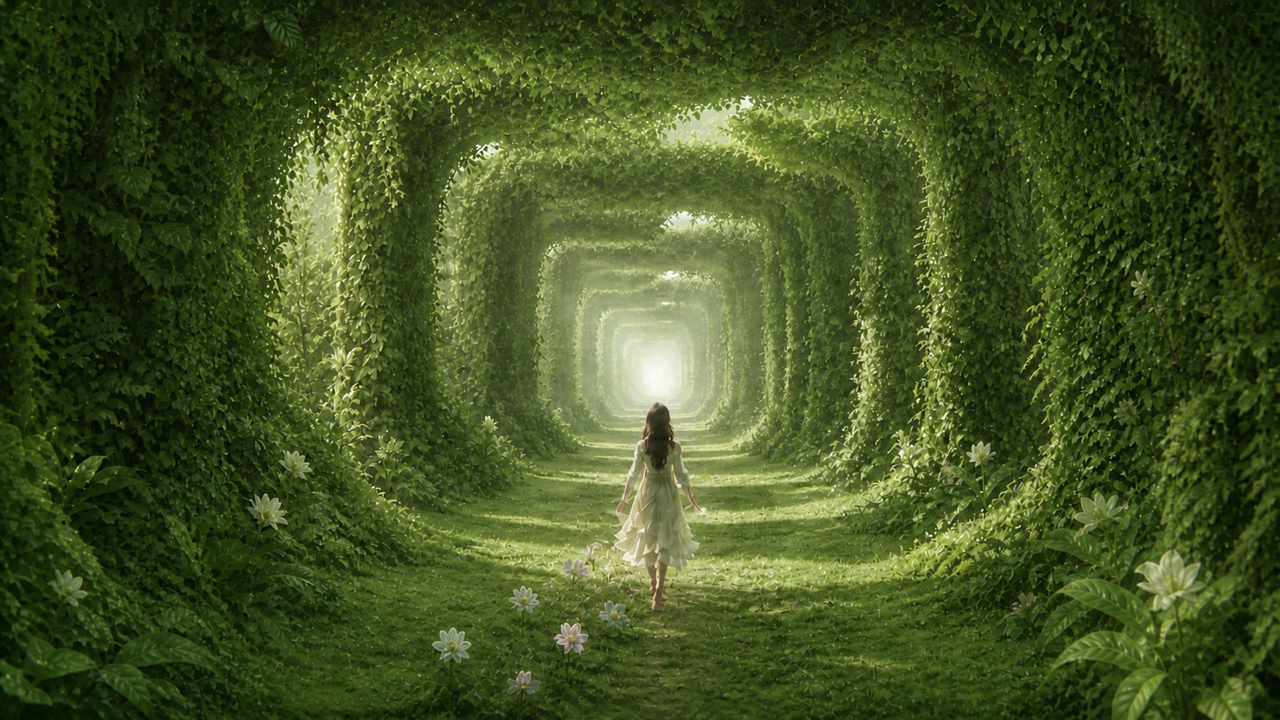}\hspace{0.8mm}%
  \teasertile{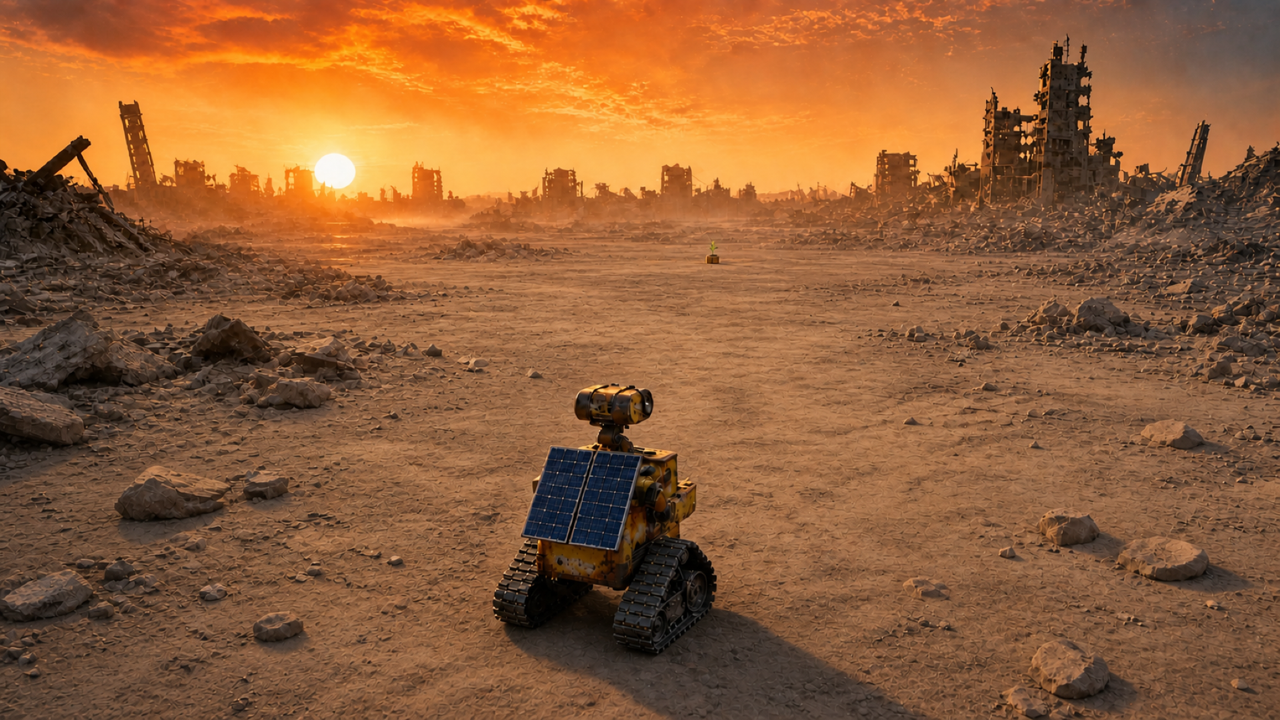}\hspace{0.8mm}%
  \teasertile{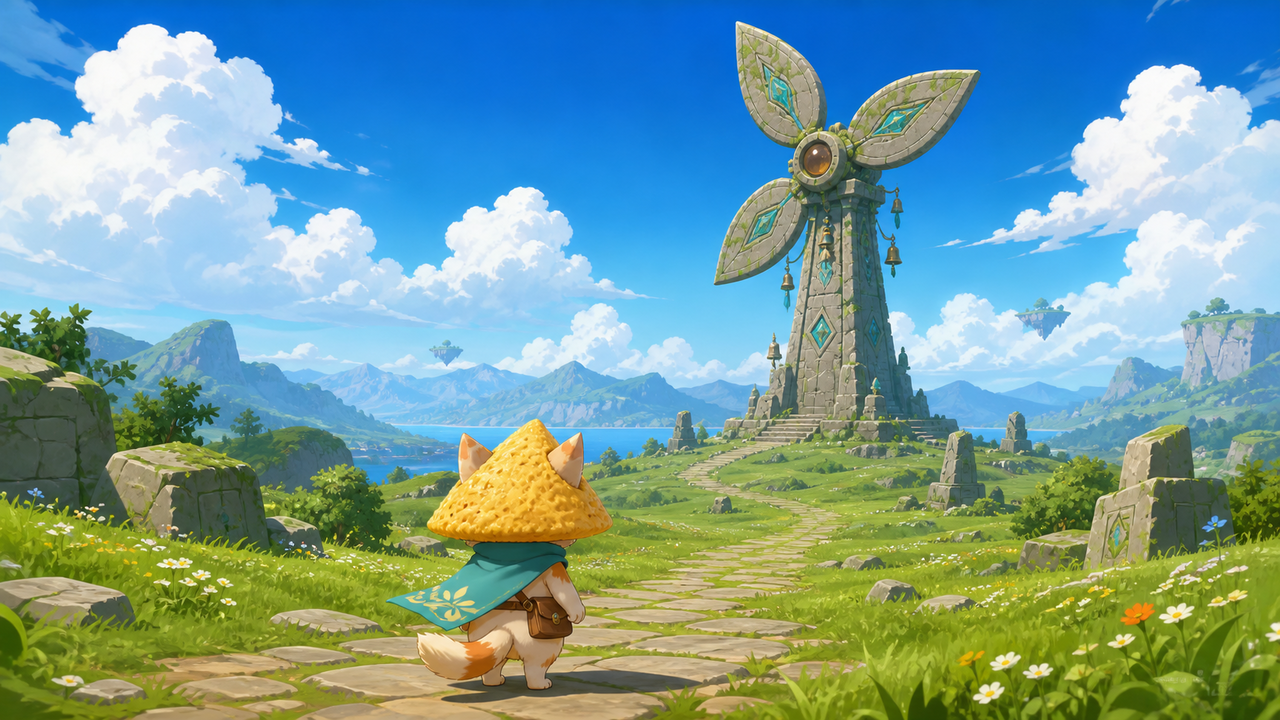}\hspace{0.8mm}%
  \teasertile{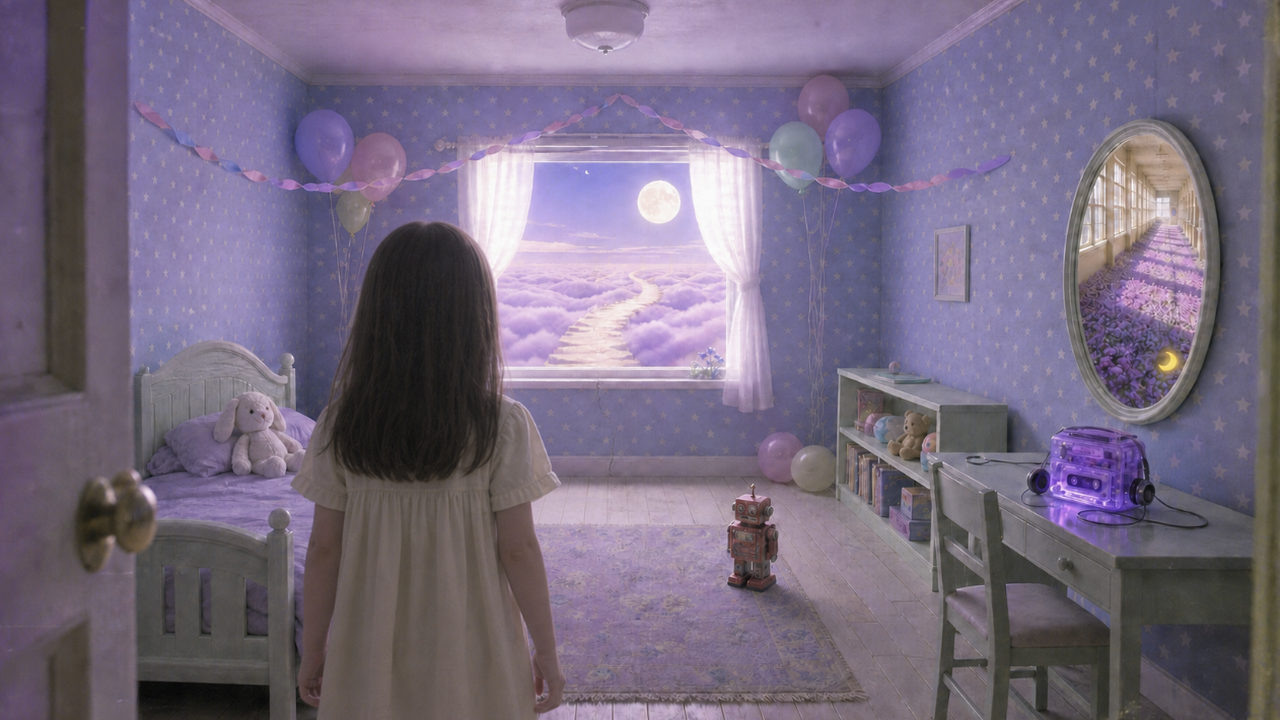}%
}
\endgroup
\captionsetup{hypcap=false}
\captionof{figure}{Diverse environments and visual styles in Zing-0.5.}
\label{fig:teaser}
\end{minipage}\par
\endgroup

\clearpage

\section{Introduction}
\label{sec:introduction}

The Zing series is designed around playability. Users should be able to explore generated worlds, influence what happens, and use the world's responses to decide what to do next. Exploration offers discovery and novelty, but movement and observation alone give users limited ways to create situations or pursue their intentions. We therefore aim to support intervention in events and behavior, so that discoveries lead to actions and feedback guides further attempts.

Interactive world models have advanced keyboard and camera control~\citep{gamengen,matrix2,lingbot,worldplay,abotworld}, while language-conditioned systems extend interaction to changes in content and behavior~\citep{gamegenx,gamecraft2,infinity}. These controls serve complementary purposes: keyboard inputs provide direct, continuous movement and view control, while text expresses intended changes to characters, events, or the surrounding environment. Joint control lets users form intentions while exploring, intervene in the current situation through language, and continue moving to observe the result. Connecting exploration, intervention, and feedback within one ongoing session is our starting point for playable generated worlds.

\begin{figure}[H]
  \centering
  \includegraphics[width=\linewidth]{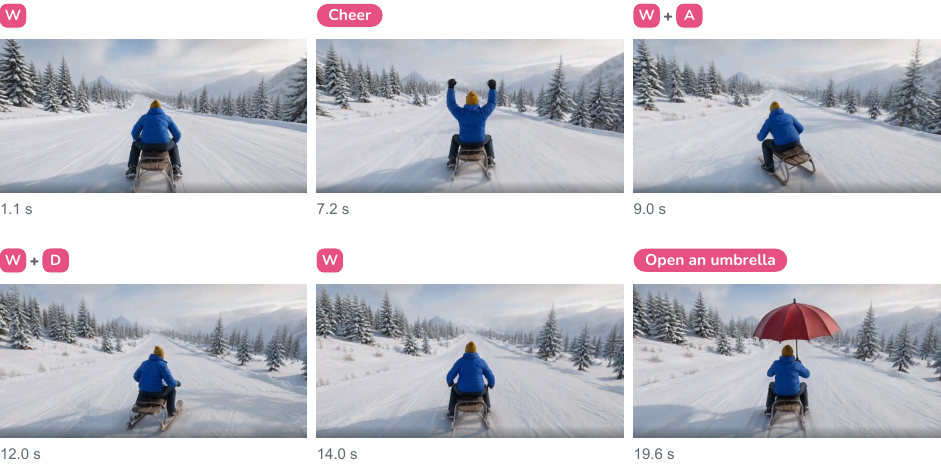}
  \caption{\textbf{Joint action and text control.} Six frames from one interactive session show a sled rider navigating a snowy landscape while responding to text instructions to cheer and open an umbrella. Key symbols summarize movement directions, while text labels summarize event instructions. Timestamps refer to the recording.}
  \label{fig:jointdemo}
\end{figure}

We introduce \mbox{Zing-0.5} (\figref{fig:teaser}), a 5B autoregressive world model built on Wan2.2-TI2V-5B~\citep{wan,wan22}. It combines keyboard navigation with online text instructions and continues generation from the existing visual context. Jointly annotated videos align actions and text with their visual outcomes, while continuous action strengths provide fine-grained movement and view control. In \figref{fig:jointdemo}, the user navigates a sled through a snowy landscape and instructs the rider to cheer and open an umbrella, all within the same generation session.

This interaction requires reconciling the temporal scales of event learning and incremental generation. A text-directed event can span multiple generation blocks, while ongoing interaction requires short continuations that can incorporate new user input. We train a segment-level teacher on connected multi-prompt videos to jointly model each prompt segment and learn continuation across prompt changes. Distribution-matching distillation then uses this teacher to supervise a block-level causal student, providing event supervision that spans multiple student blocks. We further train the student on its own generated histories using rollout and replay~\citep{sgf}, and use long-sequence adaptation and history perturbation to prepare it for extended interaction.

To support sustained exploration and repeated attempts, interaction must run in real time at an affordable cost. Four-step generation and a streaming runtime support 480p inference at 24 FPS at an estimated server rental cost of approximately \$\zingStreamMinuteCost{} per stream-minute. The runtime retains visual context across prompt updates and combines cache reuse, lightweight local decoding, and bounded media queues. Section~\ref{sec:inference-system} describes the deployment and streaming pipeline.

On WBench Navigation~\citep{wbench}, \mbox{Zing-0.5} achieves an overall score of 81.0 and a consistency score of 88.5 across 158 image-conditioned cases. These scores evaluate navigation-conditioned generation; the session in \figref{fig:jointdemo} separately provides qualitative evidence of online text control during continued navigation. We release the model weights, inference code, and Zing-SGLang serving implementation to support further work on playable generated worlds.

\Needspace{11\baselineskip}
The report makes three contributions:
\begin{itemize}
  \item \textbf{Unified action and text conditioning.} Keyboard inputs with continuous strengths, temporally aligned text instructions, and jointly annotated videos support navigation and event control within the same sequence.
  \item \textbf{Event-scale supervision for incremental generation.} A segment-level teacher trained on multi-prompt continuations supervises a block-level causal student through distribution-matching distillation, connecting event-scale learning with incremental generation.
  \item \textbf{Low-cost real-time interaction.} Four-step generation and context-preserving streaming support 480p interaction at 24 FPS at an estimated server rental cost of approximately \$\zingStreamMinuteCost{} per stream-minute.
\end{itemize}

\section{Related Work}
\label{sec:related}

\paragraph{Interactive world models.}
Interactive world models generate a visual continuation in response to user actions, allowing users to observe the result and choose their next input. GameNGen studies this loop within a specific game environment~\citep{gamengen}, while Matrix-Game~\citep{matrix2}, LingBot-World~\citep{lingbot}, and WorldPlay~\citep{worldplay} extend controllable generation to a broader range of scenes. ABot-World-0 uses raw keyboard actions for both scene roaming and third-person character control~\citep{abotworld}. These systems establish navigation and action response as central capabilities of generated worlds.

Control interfaces differ in how they represent user intent. Camera-conditioned methods supply geometric viewpoint constraints~\citep{cameractrl,worldcam}; GameCraft~\citep{gamecraft} and WorldPlay~\citep{worldplay} connect user commands with camera or geometric representations, and JoyAI-Echo-1.5 studies a unified camera-intent interface and scale calibration~\citep{joyecho}. Zing retains directional commands and their strengths for movement and view control. Its focus is a continuous interaction session in which users can navigate and also modify the evolving world through text.

\paragraph{Language-guided world interaction.}
Language lets users specify scene changes, events, and character behavior beyond directional navigation. GameGen-X explores text-guided interactive game video generation~\citep{gamegenx}; GameCraft-2 supports instruction-driven control of camera motion, character behavior, and environment dynamics~\citep{gamecraft2}; and LingBot-World-Infinity combines diverse actions with text-driven events~\citep{infinity}. These works place semantic instructions within the interaction interface. H3-World further explores time-aligned language actions as a control representation~\citep{h3world}.

Multi-prompt video generation provides a related temporal perspective: assigning prompts to intervals or shots organizes successive events, as in Gen-L-Video~\citep{genlvideo} and CausalCine~\citep{causalcine}. An online session additionally requires accepting instructions as interaction proceeds, with future user inputs unavailable to the current prediction. Zing aligns action and text conditions on separate timelines, allowing either input to change while generation continues from the existing visual context. Jointly annotated videos provide supervision for navigation and text-directed changes within the same sequence.

\paragraph{Continuous autoregressive generation.}
Incremental generation supports ongoing interaction, but generated history can accumulate errors over time. Diffusion Forcing provides a framework for sequence generation with different noise levels across time~\citep{diffusionforcing}. History corruption, as used in GameNGen~\citep{gamengen} and Helios~\citep{helios}, exposes training to imperfect context; Self Forcing instead trains on the model's own rollouts~\citep{selfforcing}, and Self Gradient Forcing uses rollout and replay to make long-sequence training practical~\citep{sgf}. ABot-World-0's LongForcing aligns long student rollouts with an extended-horizon teacher~\citep{abotworld}.

Zing combines long-sequence adaptation, history perturbation, and generated-history training. Its teacher and student use different temporal partitions: the teacher models a prompt interval jointly, while the student produces short causal blocks. This provides supervision across the blocks over which an event unfolds. At inference, a bounded visual KV cache carries context across blocks and is retained when the text-conditioning cache is updated, supporting prompt changes within an ongoing session.

\paragraph{Efficient generation for real-time interaction.}
Few-step generation reduces the computation needed for each interactive update. CausVid combines causal generation with distillation~\citep{causvid}; consistency training~\citep{consistency} and CMT~\citep{cmt} provide routes to few-step prediction, while Causal Forcing~\citep{causalforcing} and Causal Forcing++~\citep{causalforcingpp} address causal conditioning during distillation. Distribution matching offers a complementary objective for matching generated distributions~\citep{dmd,dmd2}. Zing uses ODE initialization and local consistency training before distribution matching, incorporating guidance augmentation and data supervision following Decoupled DMD~\citep{decoupleddmd} and Data-Forcing Distillation~\citep{dfd}.

Real-time performance also depends on the execution and delivery pipeline. LongLive combines causal generation with KV caching for interactive long video~\citep{longlive}, while ABot-World-0 couples distillation with lightweight decoding and an optimized streaming stack~\citep{abotworld}. Zing combines four-step generation with cache reuse, lightweight local decoding, and bounded media queues. Section~\ref{sec:inference-system} reports the resulting deployment configurations and performance.

\section{Zing-0.5}
\label{sec:zing}

\subsection{Data Construction}
\label{sec:data}

\begin{figure}[!t]
  \centering
  \includegraphics[width=\linewidth]{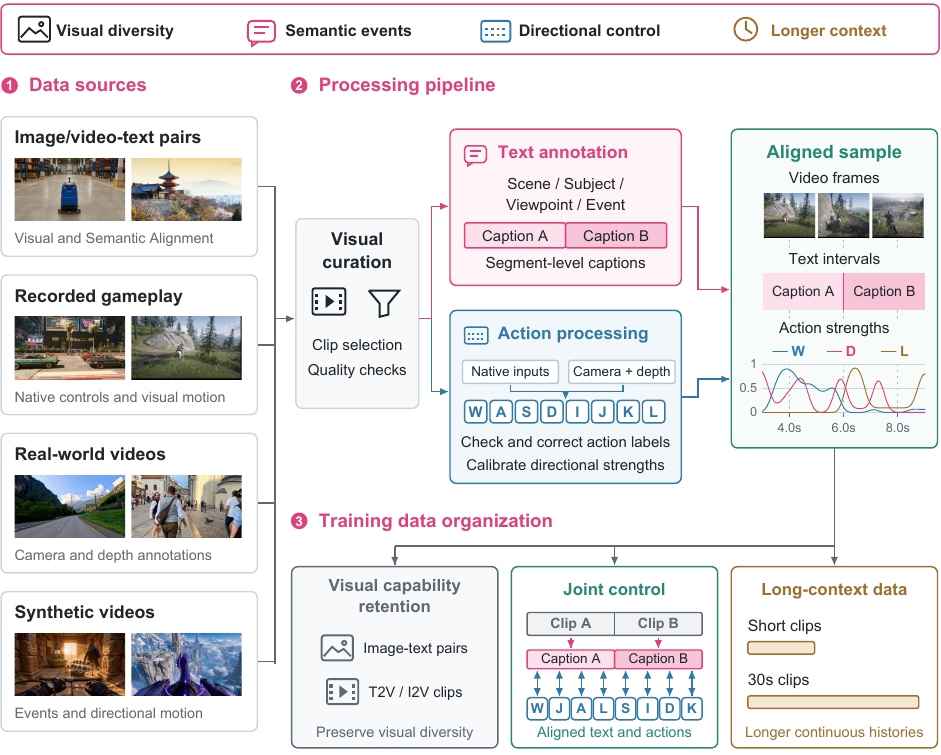}
  \caption{\textbf{Data construction.} We curate image--text and video--text pairs, gameplay, real-world videos, and synthetic videos to preserve visual capabilities and learn interactive control. Joint-control sequences align segment-level captions and continuous action strengths with video frames, while 30-second clips extend the training context.}
  \label{fig:data_construction}
\end{figure}

\textbf{Data sources.} Training an interactive world model requires broad visual coverage as well as videos that associate control inputs with changes in a scene. We combine image--text and video--text pairs, recorded gameplay, real-world videos, and synthetic videos to provide this complementary supervision~\citep{lingbot,infinity}. Image--text pairs broaden visual and semantic coverage and supervise first-frame synthesis (\secref{sec:ar}), while high-quality videos without action labels help retain the backbone's T2V and I2V capabilities. Internally collected and publicly available gameplay~\citep{abotworld} pairs recorded inputs with visual motion, while camera-annotated real-world videos~\citep{spatialvid} extend directional supervision beyond game environments. Synthetic videos supplement these sources with semantic events, directional motion, and joint-control sequences. These sources contribute different annotation types, so not every sample contains both actions and multiple text prompts. \figref{fig:data_construction} summarizes their processing and alignment.

\textbf{Visual filter.} We segment source videos into continuous clips and remove duplicate samples before annotation. Visual filtering considers brightness, sharpness, aesthetic quality, and OCR-detected text coverage to exclude poorly exposed, blurred, or text-dominated content. For subject-centric videos, we also assess subject size and framing to retain clear views of both the subject and its surroundings.

\textbf{Action annotation.} To combine these heterogeneous sources, we map their control and motion annotations to a common set of magnitude-aware directional channels (\secref{sec:architecture}). Recorded keyboard states, mouse displacements, and controller signals are converted into directional labels and strengths, retaining continuous magnitudes where available. For camera-annotated videos, we convert local camera translation and rotation into directional strengths, using depth to normalize translation. Recovered camera motion also supplements view-change labels in selected gameplay sources. We calibrate amplitudes against observed video motion to account for differences in control scale across sources. The resulting magnitudes represent relative control intensity rather than a shared physical displacement. We also rebalance the action distribution by downsampling the abundant clips containing only W-key forward motion, reducing their dominance relative to turning, view changes, and combined controls.

\textbf{Joint-control sequences.} To supervise prompt changes, we assign captions to temporal segments rather than using one description for the entire video~\citep{causalcine,infinity}. These captions describe events together with the relevant scene, subject, and viewpoint context. We retain complete two-event videos to supervise transitions and also use their extracted clips for individual-event supervision. For joint control, text intervals and action sequences are aligned to the same continuous video. Some synthetic sequences pair a text-directed event with subsequent navigation while preserving the state produced by the event. In recorded gameplay, directional inputs continue across changes in event descriptions. Both forms provide supervision for semantic events and directional control within a shared visual history.

Alongside these event-annotated clips, we include 30-second gameplay sequences to extend the motion and visual history encountered during training.

\Needspace{6\baselineskip}
\textbf{Quality control.} We refine captions and filter conspicuous mismatches between control inputs and observed motion. Action processing corrects recoverable errors in recorded controls and excludes unreliable camera-derived directions. Valid idle periods are retained, and missing action annotations remain distinct from explicitly zero-valued input.

\FloatBarrier

\subsection{Model Architecture}
\label{sec:architecture}

\begin{figure}[H]
  \centering
  \includegraphics[width=\linewidth]{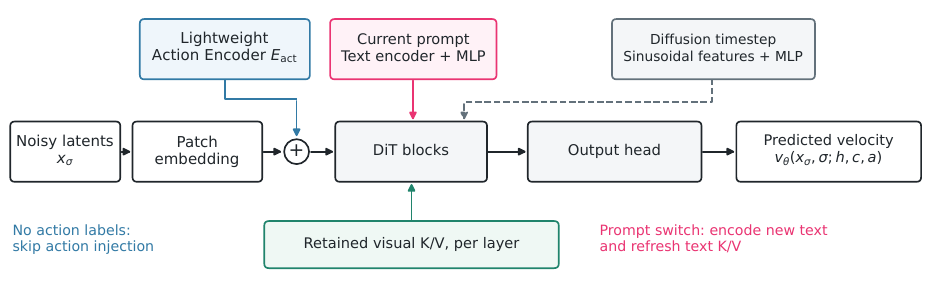}
  \caption{\textbf{Model architecture.} A lightweight action encoder adds frame-aligned control features to visual tokens before the DiT stack. Text conditions the DiT blocks through cross-attention. Prompt updates refresh text K/V while retaining visual K/V from the preceding history.}
  \label{fig:architecture}
\end{figure}

\paragraph{Problem formulation.}
Zing-0.5 builds on Wan2.2-TI2V-5B~\citep{wan22}, retaining its video autoencoder and diffusion Transformer. To enable joint action and text control, we add an action-conditioning branch and extend the existing text conditioning to support prompt changes during generation (\figref{fig:architecture}).

Under the flow-matching training objective~\citep{flowmatching}, the model predicts a velocity field from which a clean latent estimate is obtained:
\begin{equation}
  x_\sigma=(1-\sigma)x_0+\sigma\epsilon,
  \qquad
  \widehat{x}_0=x_\sigma-\sigma v_\theta(x_\sigma,\sigma;h,c,a),
  \label{eq:flow}
\end{equation}
where $x_0$ denotes clean video latents, $\epsilon\sim\mathcal{N}(0,I)$, and $\sigma\in[0,1]$. The conditioning variables $h$, $c$, and $a$ denote visual history, text, and actions. Each action feature conditions its corresponding latent frame, whereas each text prompt conditions all latent frames within its assigned temporal interval.

\paragraph{Magnitude-aware keyboard conditioning.}
\label{sec:action}

Camera trajectories provide explicit geometric constraints and are now widely used to control interactive world models~\citep{gamecraft,lingbot,worldcam}. However, camera pose reconstruction and scene-dependent scale calibration for uncalibrated video depend on the accuracy of annotation models, whose errors and outliers can destabilize training~\citep{worldplay}. Since camera poses specify states rather than user actions, a common inference scheme converts discrete keyboard inputs into camera pose increments and uses them to update the conditioning trajectory~\citep{gamecraft,worldcam}.

If the generated motion deviates from this trajectory, subsequent camera pose updates can become inconsistent with the visual history. Without feedback from generated observations, this discrepancy can compound over long rollouts, potentially causing a breakdown in visual coherence. Moreover, our requirement for joint action and prompt control adds a further complication because prompt updates may induce viewpoint changes that keyboard-based camera pose updates do not account for, increasing the risk of such discrepancies.

We therefore condition Zing-0.5 directly on keyboard inputs, avoiding an intermediate camera pose representation. This simplifies scaling data collection through precisely recorded native actions and keeps the control representation consistent between training and inference. However, encoding motion solely as discrete directional labels discards the fine-grained magnitude information available in camera pose trajectories~\citep{disco,infiniteworld}. We therefore augment each directional key with a continuous magnitude to represent motion intensity while retaining its discrete identity. These magnitudes also provide adjustable control over movement and view changes at inference. At video-frame transition $t$, the action is represented by
\begin{equation}
  a_t=(a_t^{\mathrm{W}},a_t^{\mathrm{A}},a_t^{\mathrm{S}},a_t^{\mathrm{D}},
       a_t^{\mathrm{I}},a_t^{\mathrm{J}},a_t^{\mathrm{K}},a_t^{\mathrm{L}}).
  \label{eq:action}
\end{equation}
Here, W/A/S/D encode movement and I/J/K/L encode view changes. Each component is a nonnegative strength, and multiple channels can be active simultaneously. The magnitudes represent relative control intensity rather than metric displacement. Both native recordings and directional labels derived from camera poses use this representation, with source-dependent calibration as described in \secref{sec:data}.

After video-frame sampling, we average the selected transition controls within each latent-frame window $\mathcal{W}_j$ to align them with the temporally compressed video. A causal encoder $E_{\mathrm{act}}$ maps the aligned controls to features added to the corresponding visual tokens $z_{j,p}$:
\begin{equation}
  \bar a_j=\frac{1}{|\mathcal{W}_j|}\sum_{t\in\mathcal{W}_j}a_t,
  \qquad e_j=E_{\mathrm{act}}(\bar a_{\leq j}),
  \qquad z_{j,p}\leftarrow z_{j,p}+e_j.
  \label{eq:actioninjection}
\end{equation}
Here $p$ indexes spatial positions within a latent frame. The encoder uses sinusoidal magnitude embeddings and a residual MLP, followed by causal temporal convolutions that aggregate recent control history. A projection to the model dimension produces $e_j$, which is broadcast over spatial tokens before the Transformer stack. This projection is zero-initialized to preserve the pretrained function at the start of adaptation. The action encoder is lightweight, adding only 3.68M parameters (approximately 0.074\% of the 5B backbone).

\Needspace{12\baselineskip}
\paragraph{Multi-prompt conditioning.}
\label{sec:joint}

Earlier navigation-focused world models typically pair directional inputs with a fixed global prompt~\citep{gamecraft,worldcam}, limiting text-based control over changes to subjects and scenes during generation. Other systems expose navigation and text-directed generation as separate interaction modes~\citep{happyoyster}. Zing-0.5 combines keyboard control and prompt updates within a single continuous session. We group text-driven interactions into three categories: subject changes, scene changes, and subject--scene interactions. To support these interactions, we align each prompt with its corresponding video interval rather than using Wan-2.2's video-level text conditioning: visual tokens in interval $\mathcal{S}_k$ receive frame-aligned action features and attend only to the embeddings of prompt $c_k$ through cross-attention.

During training, we extend the single-prompt supervision used in bidirectional adaptation to multi-prompt sequences in the autoregressive stage (\secref{sec:ar}). At inference, unlike LongLive's more complex KV-recache strategy~\citep{longlive}, which recomputes historical visual features under the new prompt, we update only the text K/V cache and retain the visual K/V cache, preserving scene context while incorporating the new instruction.

\FloatBarrier

\subsection{Progressive Training and Distillation}
\label{sec:training}

\begin{figure}[H]
  \centering
  \includegraphics[width=\linewidth]{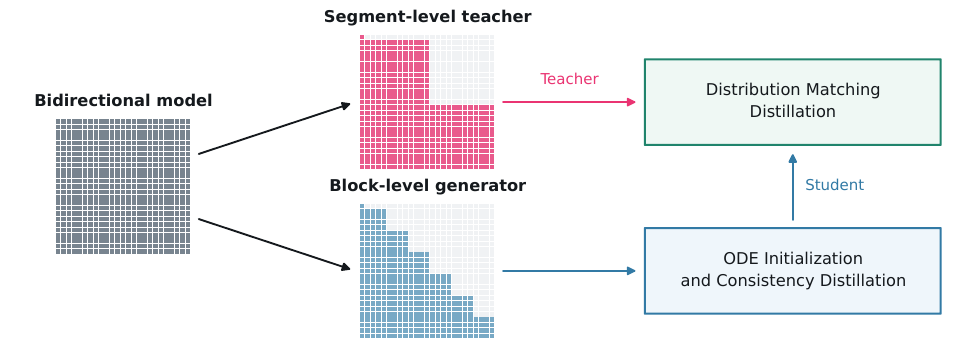}
  \caption{\textbf{Training procedure.} Bidirectional adaptation is followed by autoregressive training of a segment-level teacher and a block-level generator. The generator undergoes ODE initialization and local consistency distillation, followed by DMD supervised by the segment-level teacher. Colored mask entries indicate allowed attention between latent frames, with queries in rows and keys in columns.}
  \label{fig:training}
\end{figure}

Our four-stage training pipeline (\figref{fig:training}) transforms a pretrained bidirectional video model into a few-step autoregressive world model with joint action and text control. Bidirectional adaptation (\secref{sec:bidir}) first introduces action conditioning, with action-free image and video supervision to preserve the pretrained generation capabilities. Autoregressive adaptation (\secref{sec:ar}) then introduces prompt-transition supervision in two causal branches: a segment-level teacher for extended events and a block-level generator for incremental generation. ODE initialization and local consistency distillation (\secref{sec:odecd}) prepare the block-level generator for few-step sampling. Finally, distribution matching distillation (\secref{sec:dmd}) trains this student on its own rollouts under supervision from the segment-level teacher, addressing the shift from training-data histories to generated context.

\subsubsection{Bidirectional Adaptation}
\label{sec:bidir}

Starting from Wan2.2-TI2V-5B~\citep{wan22}, we retain the pretrained backbone and its bidirectional attention while introducing the action-conditioning branch. To preserve image and video generation capabilities while learning keyboard control, we mix action-annotated videos with high-quality, action-free T2I, T2V, and I2V samples (\secref{sec:data}). Each sample uses a global text prompt, and training follows the flow-matching objective in \eqnref{eq:flow}. After adaptation on short sequences, an additional phase with 30-second videos extends the temporal context seen during training, providing a critical foundation for long-horizon generation in the subsequent autoregressive stage.

\Needspace{17\baselineskip}
\subsubsection{Autoregressive Adaptation}
\label{sec:ar}

\begin{figure}[!t]
  \centering
  \includegraphics[width=\linewidth]{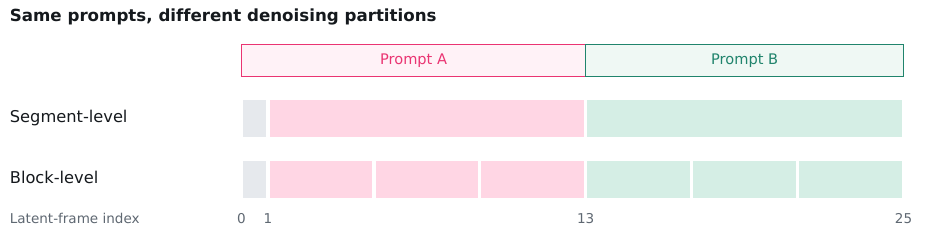}
  \caption{\textbf{Temporal partitions for autoregressive adaptation.} Rectangles denote denoising units. Both branches separate the first latent frame, then use prompt segments (teacher) or four latent frames (generator).}
  \label{fig:training-partitions}
\end{figure}

\paragraph{Segment-level teacher.}
Text-directed events often span multiple generation blocks. A block-level teacher cannot use later blocks when supervising earlier parts of the same event. The teacher is used during distillation rather than online generation, so it need not share the generator's latency constraint. We therefore train a segment-level teacher that jointly denoises each prompt interval with bidirectional attention (\figref{fig:training-partitions}). This retains the pretrained model's joint temporal modeling and allows supervision to incorporate context from the entire interval. To learn continuation across prompt changes, we train on connected multi-prompt videos with causal attention between segments, conditioning each segment on the preceding visual history.

\paragraph{Block-level generator.}
While the teacher can jointly model an entire prompt segment, the generator must produce video incrementally to respond to user inputs with low latency. We therefore train a separate autoregressive branch from the same bidirectionally adapted model, using blocks of four latent frames rather than full prompt segments. At inference, this block-level generator produces each new block from the preceding visual history, conditioned on the current action and text inputs.

\paragraph{Separate first-frame modeling.}
We find that a low-quality initial frame makes autoregressive generation more prone to compounding errors. We therefore separate first-frame synthesis from video continuation, reformulating T2V as T2I followed by I2V. This allows us to directly supervise first-frame synthesis with a large number of high-quality image--text pairs, while T2V and I2V can share the same continuation process from a generated or supplied image. Applying this separation to both the segment-level teacher and the block-level generator makes their first-frame denoising structures consistent during distillation (\figref{fig:training-partitions}). For the block-level generator, this yields the nominal $1+4n$ latent-frame layout.

After training on short sequences, both autoregressive branches undergo an additional phase with 30-second videos to learn longer continuations.

\Needspace{8\baselineskip}
\paragraph{History augmentation.}
Teacher forcing supplies clean ground-truth history, whereas inference depends on previously generated frames. Following Helios~\citep{helios}, we augment the block-level generator's conditioning history to reduce sensitivity to imperfect generated context. Training mixes clean, noisy, and spatially blurred histories, with an additional perturbation that rescales deviations from the latent-channel mean, while the target latents and reference frames remain uncorrupted.

\subsubsection{ODE Initialization and Consistency Distillation}
\label{sec:odecd}

A model trained to predict a local flow direction need not make accurate endpoint predictions across large noise intervals. We first train that mapping with teacher ODE trajectories, following the mid-training rationale of CMT~\citep{cmt}, and then refine it with local consistency distillation~\citep{consistency}. Our initialization teacher is causal and uses ground-truth history. This differs from CausVid's bidirectional-teacher initialization~\citep{causvid} and avoids giving the teacher future context unavailable to the student, the mismatch studied by Causal Forcing~\citep{causalforcing}.

Teacher integration in both stages uses classifier-free guidance~\citep{cfg}, combining a text-and-action prediction with a negative-text, no-action prediction. Both teacher evaluations use the same history.

\Needspace{12\baselineskip}
\paragraph{ODE initialization.}
The frozen block-level teacher generates a denoising trajectory from Gaussian noise under the aligned text and actions. Intermediate states $x_{\sigma_i}^{\mathrm{T}}$ are paired with that trajectory's terminal prediction $x_0^{\mathrm{T}}$. Writing $F_\theta$ for the clean prediction in \eqnref{eq:flow}, we minimize
\begin{equation}
  \mathcal{L}_{\mathrm{ODE}}
  =\mathbb{E}\biggl[\Bigl\lVert
       F_\theta(x_{\sigma_i}^{\mathrm{T}},\sigma_i;h,c,a)-x_0^{\mathrm{T}}
     \Bigr\rVert_{M}^{2}\biggr],
  \label{eq:ode}
\end{equation}
where $M$ selects target tokens and $\|\cdot\|_M^2$ denotes their mean squared error. The target is the teacher's endpoint, not necessarily the clean latent of the training video. A separate data-endpoint regression term retains direct supervision from the data.

Trajectories are generated online and cached for reuse. Both teacher sampling and student endpoint regression use unperturbed ground-truth history. We increase the sequence-length budget during this stage before proceeding to consistency training.

\paragraph{Local consistency distillation.}
We next apply local consistency distillation~\citep{consistency}, training endpoint predictions at nearby noise levels to agree. Causal Forcing++ studies a closely related initialization for causal video generation~\citep{causalforcingpp}. From a noised data state $x_{\sigma_i}$, the frozen teacher integrates to $x_{\sigma_j}$, with $\sigma_j<\sigma_i$. An exponential-moving-average student supplies the target:
\begin{equation}
  \mathcal{L}_{\mathrm{CD}}
  =\mathbb{E}\biggl[\Bigl\lVert
      F_\theta(x_{\sigma_i},\sigma_i)
      -\operatorname{sg}\bigl(F_{\bar\theta}(x_{\sigma_j},\sigma_j)\bigr)
    \Bigr\rVert_{M,w}^2\biggr].
  \label{eq:cd}
\end{equation}
The conditioning arguments are omitted for brevity, and $\operatorname{sg}$ denotes stop-gradient. The weighted norm $\|\cdot\|_{M,w}^2$ sums weighted target-token mean squared errors and divides by the sum of their weights. We increase the weights linearly with the ending schedule index, emphasizing pairs that end closer to the clean state. When integration reaches the clean endpoint, that state is used directly as the target.

The student, moving-average target, and conditional teacher evaluation use the same positive controls. All evaluations, including the negative teacher branch, share one sampled noise-only history perturbation.

\subsubsection{Distribution Matching Distillation}
\label{sec:dmd}

We now train the student on its own autoregressive outputs, following the on-policy motivation of Self Forcing~\citep{selfforcing}. A frozen teacher estimates the target-distribution score, and a learned fake scorer estimates the distribution of the student's outputs~\citep{dmd,dmd2}. Both are initialized from the segment-level branch. They assess an extended prompt segment that the block-level student generates incrementally.

\paragraph{Rollout and replay.}
Our DMD training covers text-to-image generation, video generation with multiple prompts, and action-conditioned generation. Prompt intervals and action sequences have a prescribed temporal extent, so each rollout must follow the length and temporal structure of its corresponding ground-truth sample. A prompt paired with a globally fixed rollout duration is therefore insufficient. These heterogeneous sequence lengths also require data packing for efficient scorer evaluation and student optimization. We generate each instance independently with a KV cache, but retaining gradients through successive cache updates would make long rollouts prohibitively expensive in memory.

We adapt the two-pass procedure of Self Gradient Forcing~\citep{sgf} to separate per-instance generation from packed gradient computation. First, the student performs a no-gradient rollout for each instance, caching its history and recording noisy states at sampled denoising exits, their noise levels, and the resulting clean predictions. Next, we pack the generated samples and run the frozen real scorer and the learned fake scorer to compute detached surrogate gradients. Finally, the student replays the recorded inputs in a packed teacher-forcing pass with gradient tracking enabled, applies the stored surrogate gradients to its recomputed predictions, and updates its parameters. The generated history latents remain detached; their context representations are recomputed during replay, so gradients flow through this packed pass without extending back through the rollout cache.

This ordering allows both scorers to be offloaded during rollout. It also defers construction of the student's computation graph and backpropagation until after scoring, avoiding overlap between scorer evaluation and the student's saved activations. Together, these choices reduce peak GPU memory usage when training on long, variable-length samples.

\paragraph{Distillation objective.}
Let $x_{\mathrm{roll}}$ denote the student's clean predictions recorded during rollout. Before student replay, we renoise these predictions and obtain clean predictions from the fake scorer, $\widehat{x}_{\mathrm{f}}$, and the conditional teacher, $\widehat{x}_{\mathrm{r},\mathrm{c}}$. A second, independent renoising draw supplies the teacher predictions $\widehat{x}'_{\mathrm{r},\mathrm{c}}$ and $\widehat{x}'_{\mathrm{r},\mathrm{u}}$, with and without positive text and action conditioning. Following Decoupled DMD~\citep{decoupleddmd}, the distribution-matching and guidance-augmentation directions are
\begin{equation}
  g_{\mathrm{DM}}=
  \frac{\widehat{x}_{\mathrm{f}}-\widehat{x}_{\mathrm{r},\mathrm{c}}}{Z_{\mathrm{DM}}},
  \qquad
  g_{\mathrm{CA}}=
  \gamma\frac{\widehat{x}'_{\mathrm{r},\mathrm{u}}-\widehat{x}'_{\mathrm{r},\mathrm{c}}}{Z_{\mathrm{CA}}},
  \label{eq:dmd}
\end{equation}
where $\gamma$ controls guidance augmentation. Each $Z$ is computed separately for each sample as the mean absolute difference between $x_{\mathrm{roll}}$ and the corresponding conditional teacher prediction over its target tokens, with a lower bound to prevent division by a near-zero value. We add the two directions and weight their sum according to the DMD--DFD mixture described below, giving a detached gradient signal $g$. During the final replay pass, the student recomputes the corresponding clean predictions $x_{\mathrm{G}}$ with gradient tracking enabled. The surrogate loss $\tfrac12\bigl\lVert x_{\mathrm{G}}-\operatorname{sg}(x_{\mathrm{G}}-g)\bigr\rVert_M^2$ applies the stored signal to update the student without backpropagating through either scorer. The fake scorer is trained separately with a flow-matching loss on generated videos.

Data-Forcing Distillation (DFD)~\citep{dfd} was originally applied as a post-training stage after DMD2 training had stabilized. We incorporate DFD throughout the DMD stage by probabilistically assigning a subset of samples in each batch to DFD and using standard DMD for the remaining samples. For DFD samples, the teacher receives noised ground-truth latents and ground-truth history, while the fake scorer continues to evaluate generated samples. We weight the contributions of DFD and DMD samples to smooth the combined gradient update.

Empirically, this joint training reduces fluctuations in motion magnitude and mitigates severe visual degradation. With prolonged training, however, we observe some high-frequency noise artifacts and reduced color saturation.

The final student uses four denoising steps per block. Guidance is incorporated into training, so inference does not require a second, unconditional forward pass.

\section{System Infrastructure}
\label{sec:inference-system}

\subsection{Training System}
\label{sec:infra}

Zing-0.5 uses a token-budgeted, variable-length input pipeline so images, short clips, and long sequences can share a batch without padding to the longest sample. A global coordinator samples data sources and advances deterministic shuffling; CPU workers decode, resize, and align prompts and actions; and GPU workers perform cache-backed VAE encoding. The resulting packed tensors carry visual tokens together with \texttt{seq\_lens}, block identifiers, 3D positions, target masks, prompt spans, and token-aligned actions, preserving each sample's boundaries.

The 5B backbone consumes this contract directly. Bidirectional adaptation uses variable-length FlashAttention, while causal teacher forcing uses a sparse FlexAttention block mask and never materializes a dense token-by-token mask. For long sequences, one worker broadcasts each pack within its Ulysses sequence-parallel group; attention exchanges sequence and head partitions through all-to-all communication, and FSDP shards parameters and reduces the partial gradients. Activation checkpointing limits memory, while cached visual latents and text embeddings avoid repeated encoding.

\subsection{Low-Cost Real-Time Inference on RTX 5090}
\label{sec:realtime-inference}

On a server with \zingRtxGpuCount{} RTX 5090 GPUs, Zing-0.5 serves \zingConcurrentStreams{} independent \zingRtxResolution{} streams at a client-visible \zingPlaybackFPS{} FPS. The unpaced steady-state measurement reaches \zingMeasuredFPS{} FPS, leaving operating margin above the real-time playback rate. Each GPU hosts a complete replica: the DiT, its bounded causal KV cache, and a local TAEHV decoder. This one-stream-per-GPU layout avoids cross-GPU latent transfer and keeps every stream self-contained on one GPU.

A regular causal block uses four denoising steps to generate \zingLatentFramesPerBlock{} latent frames, which decode to \zingDecodedFramesPerBlock{} video frames. The 5090 path keeps the DiT and recurrent decoder resident, selects FA4 for SM120, and reuses rotated keys, RoPE tensors, packed-attention metadata, and action conditioning. The native VAE is needed only for initial-image encoding; recurrent blocks use the much lighter TAEHV decoder.

After local decoding, frames enter the media path directly instead of returning through an intermediate RGB service. H.264/fMP4 fragments are placed in a non-blocking bounded ring that favors the live edge under backpressure. Browser demuxing and decoding run in a worker behind a short playback queue, while ordered completion markers and an end-of-stream barrier preserve session correctness.

\paragraph{Context management.}
Full-history attention would increase memory use and per-block computation as an interaction continues. Following LongLive-2.0~\citep{longlive2}, we maintain a bounded visual KV cache with a fixed prefix sink and a sliding window of recent context. The sink keeps the initial visual context accessible throughout generation, while the window advances with each new block and discards older, unprotected entries. At a prompt update, we replace the text K/V cache while retaining the visual K/V cache, as described in \secref{sec:architecture}. To retain a visual reference under the new instruction, the streaming runtime marks the first latent frame generated after the switch and pins its K/V entries when they would leave the regular window. A newly activated pin replaces the previous one and uses part of the recent-context budget. The sink, pin, and sliding window thus share a fixed capacity, keeping visual KV storage bounded across repeated prompt changes.

\section{Results}
\label{sec:results}

\subsection{Quantitative Results}
\label{sec:quantitative}

\paragraph{WBench Navigation.}
We evaluate \mbox{Zing-0.5} on the 158-case Navigation split of WBench~\citep{wbench}, using the provided initial images and prescribed action controls. Videos are generated at $1248\times704$ resolution with four denoising steps per block and saved at 24 frames per second. We follow the official Navigation evaluation and aggregation across quality, setting, interaction, consistency, and physical plausibility. With only 5B parameters, \mbox{Zing-0.5} achieves a competitive overall score of 81.0 (\tabref{tab:wbench}).

\begin{table}[htbp]
  \centering
  \small
  \setlength{\tabcolsep}{4.4pt}
  \caption{\textbf{WBench Navigation, 158 cases.} Selected official leaderboard results as of September 9, 2026~\citep{wbenchboard}. Scores use the benchmark's 0--100 scale; higher is better.}
  \label{tab:wbench}
  \begin{tabular}{lrrrrrr}
    \toprule
    \textbf{Model} & \textbf{Avg.} & \textbf{Quality} & \textbf{Setting} & \textbf{Interact.} & \textbf{Consist.} & \textbf{Physical} \\
    \midrule
    JoyAI-Echo-1.5, bidirectional~\citep{joyecho} & \textbf{81.6} & 81.5 & 79.4 & 86.6 & \textbf{89.8} & 70.6 \\
    JoyAI-Echo-1.5, 4-step~\citep{joyecho} & 81.0 & 81.1 & 77.5 & \textbf{87.9} & 88.3 & 70.1 \\
    \textbf{\mbox{Zing-0.5}, 5B, 4-step} & 81.0 & 80.6 & 77.8 & 84.2 & 88.5 & \textbf{73.8} \\
    HiDream-O1-World & 80.9 & 81.0 & 82.2 & 80.0 & 88.0 & 73.3 \\
    Alaya-EVOKE, 3-step~\citep{evoke} & 80.8 & \textbf{82.8} & \textbf{83.8} & 78.6 & 86.9 & 72.1 \\
    LingBot-World v2, fast~\citep{infinity} & 79.4 & 81.8 & 76.8 & 82.8 & 86.5 & 69.1 \\
    \bottomrule
  \end{tabular}
\end{table}

\subsection{Qualitative Results}
\label{sec:qualitative}

We complement the quantitative evaluation with recorded sessions showing how users can explore and influence generated worlds. Our focus is on visual continuity as keyboard navigation and text-directed events shape the ongoing experience.

\begin{figure}[H]
  \centering
  \includegraphics[width=\linewidth]{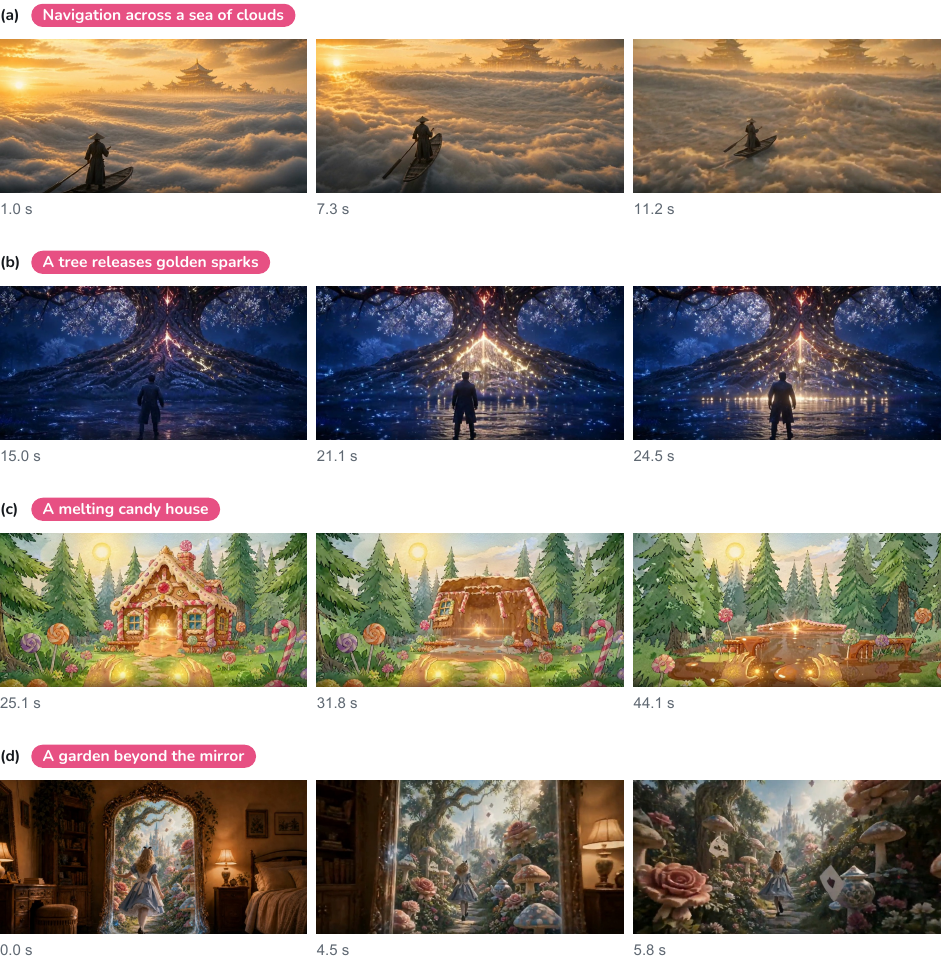}
  \caption{\textbf{Navigation, scene dynamics, and exploration.} The sequences show (a) keyboard navigation across a sea of clouds, (b) a tree releasing golden sparks after a text instruction, (c) first-person interaction with a melting candy house, and (d) passage through a mirror into a garden. Each row follows a single session, with timestamps referring to its recording.}
  \label{fig:qualitative-interactions}
\end{figure}

\paragraph{Keyboard-controlled navigation.}
In \figpanelref{fig:qualitative-interactions}{a}, the user guides a boat across a sea of clouds through keyboard movement and view controls. The boat advances toward distant palaces as the viewpoint changes, retaining the character and the surrounding sunset scene.

\paragraph{Text-conditioned scene dynamics.}
Text instructions extend control from navigation to events within the scene. In \figpanelref{fig:qualitative-interactions}{b}, an instruction to ignite a silver-flowered tree is followed by intensified light and a shower of golden sparks. The event develops within the existing forest clearing, with the tree and surrounding landscape providing a consistent setting for the change.

\paragraph{Subject--environment interaction.}
Changes to the environment can also involve the subject acting on an existing object. \figpanelref{fig:qualitative-interactions}{c} depicts first-person hands directed toward a candy house as it melts. The house loses its structure while the hands and surrounding forest remain in view.

\paragraph{Cross-scene exploration.}
Exploration can extend beyond the initial environment. In \figpanelref{fig:qualitative-interactions}{d}, the viewpoint follows a character through a bedroom mirror into a garden and continues along an outdoor path.

\begin{figure}[H]
  \centering
  \includegraphics[width=\linewidth]{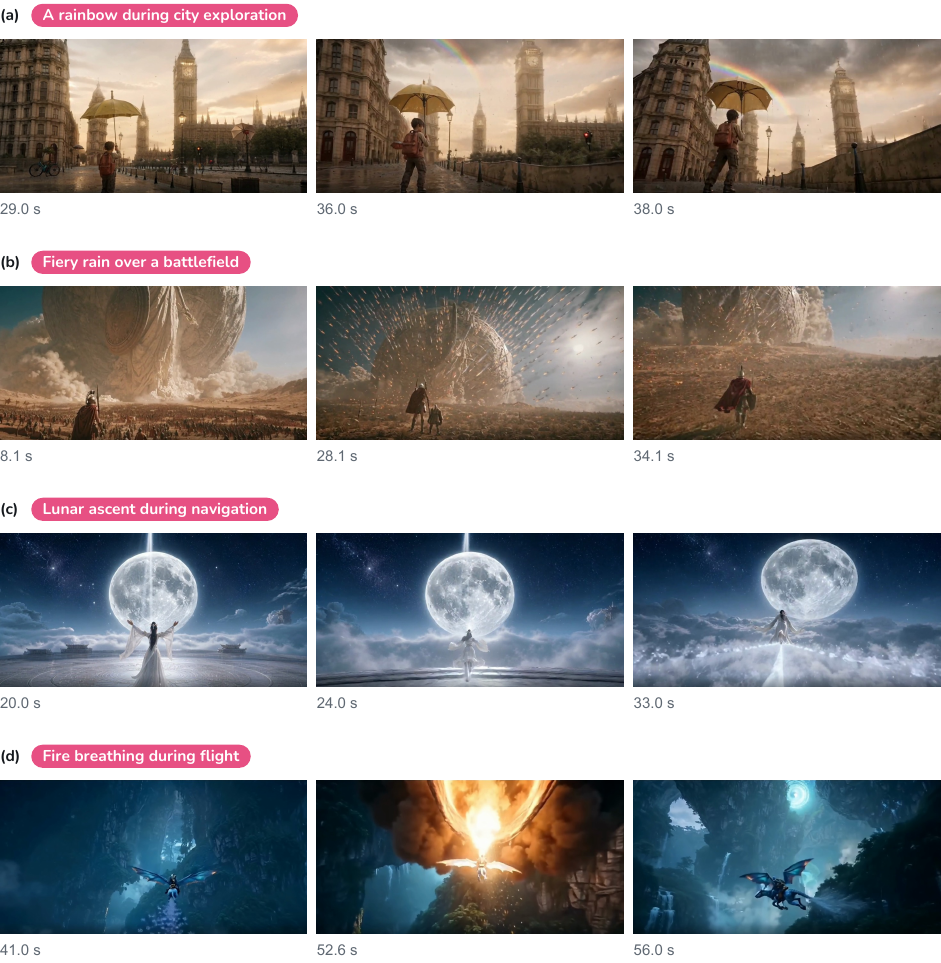}
  \caption{\textbf{Joint action and text control.} Online text instructions introduce (a) a rainbow over a city and (b) fiery rain over a battlefield, or direct (c) a character to rise into flight and (d) a flying mount to breathe fire. Each session combines these environmental or subject-directed changes with keyboard navigation. Timestamps refer to each recording.}
  \label{fig:qualitative-continuation}
\end{figure}

\paragraph{Joint action and text control.}
\figref{fig:qualitative-continuation} combines keyboard navigation with text-directed changes to the environment or subject. In \hyperref[fig:qualitative-continuation]{(a)} and \hyperref[fig:qualitative-continuation]{(b)}, online instructions introduce a rainbow and a shower of fiery projectiles, respectively, while keyboard inputs guide exploration.

For subject-directed changes, the character rises into flight following an invocation of lunar light in \figpanelref{fig:qualitative-continuation}{c}, while a mount breathes fire during flight in \figpanelref{fig:qualitative-continuation}{d}. Both behaviors unfold within keyboard-controlled sessions.

\FloatBarrier

\section{Discussion and Limitations}
\label{sec:limitations}

\paragraph{Joint Control for Playable Worlds.}
We regard playability as a central requirement for interactive worlds designed for human participation. Users need ways to act on the environment, receive meaningful feedback, and decide what to do next. Navigation supports exploration, but directional inputs express only a limited range of intentions. Language extends that range to events, character behavior, and changes in the surroundings. Combining these controls lets users influence what happens while continuing to explore. \mbox{Zing-0.5} supports this interaction in a continuous video stream: in \figref{fig:jointdemo}, the user changes the sled rider's behavior through text while continuing keyboard navigation. The significance of joint control lies in this continuous exchange. Users can respond to what they see with both movement and new instructions, making generation an experience they help shape as it unfolds.

\paragraph{Meaningful Interaction Requires Lasting Consequences.}
The examples in \secref{sec:qualitative} show visible responses within selected intervals, but do not establish whether their consequences persist through later interactions or changes in viewpoint. Sustained interaction requires an action's consequences to remain valid after the immediate response. An object moved aside should remain displaced when the user returns; a rule learned through one interaction should still apply to the next. Earlier outcomes become the basis for later decisions. If those outcomes change arbitrarily, users cannot learn how the environment works or pursue goals within it. Persistent consequences are therefore part of the interaction itself. They allow individual responses to accumulate into a coherent experience.

\paragraph{Visual History Does Not Fully Specify World State.}
\mbox{Zing-0.5} generates each continuation from retained visual context, text, and directional actions. It does not separately track entity states or enforce rules governing their transitions. Information needed for later interaction must remain implicit in these representations or be recoverable from the available conditions. This creates a structural difficulty: video shows the world from particular viewpoints, while many relevant facts remain valid outside those views. An object can leave the frame without ceasing to exist, and an event can end while its consequences still affect what should happen next. The bounded KV cache limits access to earlier visual evidence. Autoregressive generation adds another source of instability, since an error in one continuation can become context for subsequent predictions.

Long-sequence training and history perturbation expose the model to longer histories and imperfect context. These procedures address difficulties in continuing video, but they do not introduce an explicit mechanism for maintaining persistent facts. Likewise, the flow-matching and distillation objectives do not explicitly enforce state transitions or interaction rules. A continuation can consequently be plausible in its immediate visual context while conflicting with an earlier event. The challenge is to preserve the information that makes past actions binding on future outcomes, including when that information is no longer directly visible.

\paragraph{Persistent Worlds Need Architectural Support.}
Future architectures should treat this persistence as a design objective. They need to retain relevant facts across changes in viewpoint and instructions, update them consistently as users intervene, and keep generated outcomes aligned with what has already happened. Greater model capacity and longer context can contribute, but their value should be assessed through the reliability of interaction as well as video quality. Evaluation should ask whether users can change something, leave it behind, and later continue from the consequences of that change. This requirement connects architecture to playability: reliable consequences let users develop expectations, make plans, and build on earlier choices. Joint control gives users more influence over generation. Preserving the consequences of that influence is the next step toward sustained, meaningful participation.

\section{Conclusion}

We presented \mbox{Zing-0.5}, a 5B autoregressive world model designed for playability through real-time joint action and text control. Magnitude-aware keyboard inputs and online text instructions let users explore generated worlds, influence events, and act on the resulting feedback within one continuous stream. A segment-level teacher trained on connected multi-prompt videos supervises a block-level causal student, allowing event learning to span multiple blocks while generation remains incremental.

Four-step generation and context-preserving streaming support 480p inference at 24 FPS at an estimated server rental cost of approximately \$\zingStreamMinuteCost{} per stream-minute. Across 158 WBench Navigation cases, \mbox{Zing-0.5} achieves 81.0 overall and 88.5 for consistency. The joint-control demonstration separately shows a text-directed event change during continued navigation without restarting generation. We release the model weights, inference code, and Zing-SGLang serving implementation to support further work on playable generated worlds.

Joint control gives users more ways to shape what happens. For sustained play, the consequences of their actions must also remain reliable across later interactions. State and rule persistence remain open challenges: future architectures should keep generated outcomes consistent with earlier actions, so that users can learn from feedback, form expectations, and build on what they have already done.

\clearpage
\section*{Contributions and Acknowledgments}

\mbox{Zing-0.5} was developed by the SeedLeap.ai team. We thank the Wan team, the authors of the training and distillation methods used in this work, and the WBench team for making their research and implementations available.

We are especially grateful to Loopit's engineering, product, and operations teams for their substantial support in preparing and launching \mbox{Zing-0.5}, including improving the user experience, showcasing model capabilities, and selecting demos.

\subsection*{Contributors}

Contributors are listed alphabetically by surname, not in order of contribution.

Mingyang~Chen, Shengdong~Chen, Xiaoxiao~Fu, Bosheng~Gong, Haoyuan~Guo, Bowen~Li, Jiawen~Li, Kejun~Li, Tianpeng~Li, Yin~Liu, Haoze~Sun, Zeyang~Tian, Meng~Wang, Xinmiao~Wu, Jiangqiao~Yan, Zining~Zhao.

\clearpage
\begingroup
\small
\setlength{\bibsep}{4pt}
\bibliographystyle{abbrvnat}
\bibliography{refs}

\begin{thebibliography}{41}
\providecommand{\natexlab}[1]{#1}
\providecommand{\url}[1]{\texttt{#1}}
\expandafter\ifx\csname urlstyle\endcsname\relax
  \providecommand{\doi}[1]{doi: #1}\else
  \providecommand{\doi}{doi: \begingroup \urlstyle{rm}\Url}\fi

\bibitem[{Alibaba Cloud Global}(2026)]{happyoyster}
{Alibaba Cloud Global}.
\newblock {HappyOyster Launch Announcement}.
\newblock Official announcement, 2026.
\newblock URL
  \url{https://www.linkedin.com/posts/alibabacloudglobal_alibabaath-alibabaai-generativeai-activity-7452284094161371136-iL6g}.

\bibitem[Che et~al.(2024)Che, He, Liu, Jin, and Chen]{gamegenx}
H.~Che, X.~He, Q.~Liu, C.~Jin, and H.~Chen.
\newblock {GameGen-X: Interactive Open-world Game Video Generation}.
\newblock \emph{arXiv preprint
  \href{https://arxiv.org/abs/2411.00769v1}{arXiv:2411.00769}}, 2024.

\bibitem[Chen et~al.(2024)Chen, Monso, Du, Simchowitz, Tedrake, and
  Sitzmann]{diffusionforcing}
B.~Chen, D.~M. Monso, Y.~Du, M.~Simchowitz, R.~Tedrake, and V.~Sitzmann.
\newblock {Diffusion Forcing: Next-token Prediction Meets Full-Sequence
  Diffusion}.
\newblock \emph{arXiv preprint
  \href{https://arxiv.org/abs/2407.01392v3}{arXiv:2407.01392}}, 2024.

\bibitem[Chen et~al.(2026{\natexlab{a}})Chen, Wang, Lin, Yang, and
  Jin]{h3world}
D.~Chen, Z.~Wang, Z.~Lin, X.~Yang, and Y.~Jin.
\newblock {H3-World: Turning Language Understanding into World Control}.
\newblock \emph{arXiv preprint
  \href{https://arxiv.org/abs/2609.01560v1}{arXiv:2609.01560}},
  2026{\natexlab{a}}.

\bibitem[Chen et~al.(2026{\natexlab{b}})Chen, Liu, Jia, Wang, Ling, Qu, and
  Gao]{dfd}
S.~Chen, S.~Liu, Y.~Jia, Z.~Wang, H.~Ling, Q.~Qu, and J.~Gao.
\newblock {Data-Forcing Distillation: Restoring Diversity and Fidelity in
  Few-Step Video Generation}.
\newblock \emph{arXiv preprint
  \href{https://arxiv.org/abs/2606.18478v1}{arXiv:2606.18478}},
  2026{\natexlab{b}}.

\bibitem[Chen et~al.(2026{\natexlab{c}})Chen, Wang, Huang, Yang, Zhang, Xiao,
  Chu, Mao, Hu, Liu, Zhao, Mao, Chen, Xie, Qi, and Han]{longlive2}
Y.~Chen, L.~Wang, W.~Huang, S.~Yang, B.~Zhang, Y.~Xiao, R.~Chu, W.~Mao, Q.~Hu,
  S.~Liu, Y.~Zhao, H.~Mao, Y.-C. Chen, E.~Xie, X.~Qi, and S.~Han.
\newblock {LongLive-2.0: An NVFP4 Parallel Infrastructure for Long Video
  Generation}.
\newblock \emph{arXiv preprint
  \href{https://arxiv.org/abs/2605.18739v1}{arXiv:2605.18739}},
  2026{\natexlab{c}}.

\bibitem[Duan et~al.(2026)Duan, Huang, Jin, Li, Li, Li, et~al.]{joyecho}
N.~Duan, H.~Huang, W.~Jin, H.~Li, Y.~Li, Y.~Li, et~al.
\newblock {Long-Horizon Audio-Visual Generation for Persistent Stories and
  Interactive Worlds}.
\newblock \emph{arXiv preprint
  \href{https://arxiv.org/abs/2608.23383v2}{arXiv:2608.23383}}, 2026.

\bibitem[Gao et~al.(2026)Gao, Wang, Zhu, Chen, Liu, Bai, et~al.]{infinity}
Z.~Gao, Q.~Wang, J.~Zhu, J.~Chen, Z.~Liu, Q.~Bai, et~al.
\newblock {Infinite Worlds with Versatile Interactions}.
\newblock \emph{arXiv preprint
  \href{https://arxiv.org/abs/2607.07534v1}{arXiv:2607.07534}}, 2026.

\bibitem[He et~al.(2024)He, Xu, Guo, Wetzstein, Dai, Li, and Yang]{cameractrl}
H.~He, Y.~Xu, Y.~Guo, G.~Wetzstein, B.~Dai, H.~Li, and C.~Yang.
\newblock {CameraCtrl: Enabling Camera Control for Text-to-Video Generation}.
\newblock \emph{arXiv preprint
  \href{https://arxiv.org/abs/2404.02101v1}{arXiv:2404.02101}}, 2024.

\bibitem[He et~al.(2025)He, Peng, Liu, Wang, Zhang, Cui, et~al.]{matrix2}
X.~He, C.~Peng, Z.~Liu, B.~Wang, Y.~Zhang, Q.~Cui, et~al.
\newblock {Matrix-Game 2.0: An Open-Source, Real-Time, and Streaming
  Interactive World Model}.
\newblock \emph{arXiv preprint
  \href{https://arxiv.org/abs/2508.13009v1}{arXiv:2508.13009}}, 2025.

\bibitem[Ho and Salimans(2022)]{cfg}
J.~Ho and T.~Salimans.
\newblock {Classifier-Free Diffusion Guidance}.
\newblock \emph{arXiv preprint
  \href{https://arxiv.org/abs/2207.12598v1}{arXiv:2207.12598}}, 2022.

\bibitem[Hu et~al.(2025)Hu, Lai, Mitsufuji, and Ermon]{cmt}
Z.~Hu, C.-H. Lai, Y.~Mitsufuji, and S.~Ermon.
\newblock {CMT: Mid-Training for Efficient Learning of Consistency, Mean Flow,
  and Flow Map Models}.
\newblock \emph{arXiv preprint
  \href{https://arxiv.org/abs/2509.24526v2}{arXiv:2509.24526}}, 2025.

\bibitem[Huang et~al.(2026)Huang, Wang, Li, Jiang, and Wu]{disco}
H.~Huang, J.~Wang, Q.~Li, Y.-G. Jiang, and Z.~Wu.
\newblock {DisCo: World Models with Discrete Camera Motion Control}.
\newblock \emph{arXiv preprint
  \href{https://arxiv.org/abs/2606.07967v1}{arXiv:2606.07967}}, 2026.

\bibitem[Huang et~al.(2025)Huang, Li, He, Zhou, and Shechtman]{selfforcing}
X.~Huang, Z.~Li, G.~He, M.~Zhou, and E.~Shechtman.
\newblock {Self Forcing: Bridging the Train-Test Gap in Autoregressive Video
  Diffusion}.
\newblock \emph{arXiv preprint
  \href{https://arxiv.org/abs/2506.08009v1}{arXiv:2506.08009}}, 2025.

\bibitem[Jiang et~al.(2026)Jiang, Sun, Wang, Zhu, Wang, Zhang,
  et~al.]{abotworld}
F.~Jiang, Z.~Sun, M.~Wang, Z.~Zhu, C.~Wang, Y.~Zhang, et~al.
\newblock {ABot-World-0: Infinite Interactive World Rollout on a Single Desktop
  GPU}.
\newblock \emph{arXiv preprint
  \href{https://arxiv.org/abs/2607.19191v1}{arXiv:2607.19191}}, 2026.

\bibitem[Li et~al.(2025)Li, Tang, Xu, Wu, Zhou, Shao, et~al.]{gamecraft}
J.~Li, J.~Tang, Z.~Xu, L.~Wu, Y.~Zhou, S.~Shao, et~al.
\newblock {Hunyuan-GameCraft: High-dynamic Interactive Game Video Generation
  with Hybrid History Condition}.
\newblock \emph{arXiv preprint
  \href{https://arxiv.org/abs/2506.17201v1}{arXiv:2506.17201}}, 2025.

\bibitem[Lipman et~al.(2022)Lipman, Chen, Ben-Hamu, Nickel, and
  Le]{flowmatching}
Y.~Lipman, R.~T.~Q. Chen, H.~Ben-Hamu, M.~Nickel, and M.~Le.
\newblock {Flow Matching for Generative Modeling}.
\newblock \emph{arXiv preprint
  \href{https://arxiv.org/abs/2210.02747v2}{arXiv:2210.02747}}, 2022.

\bibitem[Liu et~al.(2025)Liu, Gao, Liu, Du, Li, Wu, et~al.]{decoupleddmd}
D.~Liu, P.~Gao, D.~Liu, R.~Du, Z.~Li, Q.~Wu, et~al.
\newblock {Decoupled DMD: CFG Augmentation as the Spear, Distribution Matching
  as the Shield}.
\newblock \emph{arXiv preprint
  \href{https://arxiv.org/abs/2511.22677v1}{arXiv:2511.22677}}, 2025.

\bibitem[Meng et~al.(2026)Meng, Liu, Ouyang, Wang, Cheng, Yu,
  et~al.]{causalcine}
Y.~Meng, Z.~Liu, H.~Ouyang, Q.~Wang, K.~L. Cheng, Y.~Yu, et~al.
\newblock {CausalCine: Real-Time Autoregressive Generation for Multi-Shot Video
  Narratives}.
\newblock \emph{arXiv preprint
  \href{https://arxiv.org/abs/2605.12496v1}{arXiv:2605.12496}}, 2026.

\bibitem[Nam et~al.(2026)Nam, Hong, Huang, Liu, Lee, Kim, et~al.]{worldcam}
J.~Nam, Y.~Hong, C.-H.~P. Huang, F.~Liu, J.~Lee, J.~Kim, et~al.
\newblock {WorldCam: Interactive Autoregressive 3D Gaming Worlds with Camera
  Pose as a Unifying Geometric Representation}.
\newblock \emph{arXiv preprint
  \href{https://arxiv.org/abs/2603.16871v1}{arXiv:2603.16871}}, 2026.

\bibitem[{Robbyant Team} et~al.(2026){Robbyant Team}, Gao, Wang, Zeng, Zhu,
  Cheng, et~al.]{lingbot}
{Robbyant Team}, Z.~Gao, Q.~Wang, Y.~Zeng, J.~Zhu, K.~L. Cheng, et~al.
\newblock {Advancing Open-source World Models}.
\newblock \emph{arXiv preprint
  \href{https://arxiv.org/abs/2601.20540v1}{arXiv:2601.20540}}, 2026.

\bibitem[Song et~al.(2023)Song, Dhariwal, Chen, and Sutskever]{consistency}
Y.~Song, P.~Dhariwal, M.~Chen, and I.~Sutskever.
\newblock {Consistency Models}.
\newblock \emph{arXiv preprint
  \href{https://arxiv.org/abs/2303.01469v1}{arXiv:2303.01469}}, 2023.

\bibitem[Sun et~al.(2025)Sun, Zhang, Wang, Wu, Wang, Wang, et~al.]{worldplay}
W.~Sun, H.~Zhang, H.~Wang, J.~Wu, Z.~Wang, Z.~Wang, et~al.
\newblock {WorldPlay: Towards Long-Term Geometric Consistency for Real-Time
  Interactive World Modeling}.
\newblock \emph{arXiv preprint
  \href{https://arxiv.org/abs/2512.14614v1}{arXiv:2512.14614}}, 2025.

\bibitem[Tang et~al.(2025)Tang, Liu, Li, Wu, Yang, Zhao, et~al.]{gamecraft2}
J.~Tang, J.~Liu, J.~Li, L.~Wu, H.~Yang, P.~Zhao, et~al.
\newblock {Hunyuan-GameCraft-2: Instruction-following Interactive Game World
  Model}.
\newblock \emph{arXiv preprint
  \href{https://arxiv.org/abs/2511.23429v1}{arXiv:2511.23429}}, 2025.

\bibitem[Valevski et~al.(2024)Valevski, Leviathan, Arar, and
  Fruchter]{gamengen}
D.~Valevski, Y.~Leviathan, M.~Arar, and S.~Fruchter.
\newblock {Diffusion Models Are Real-Time Game Engines}.
\newblock \emph{arXiv preprint
  \href{https://arxiv.org/abs/2408.14837v2}{arXiv:2408.14837}}, 2024.

\bibitem[{Wan Team}(2025)]{wan22}
{Wan Team}.
\newblock {Wan2.2: Open and Advanced Large-Scale Video Generative Models}.
\newblock Official model repository, 2025.
\newblock URL \url{https://github.com/Wan-Video/Wan2.2}.

\bibitem[{Wan Team} et~al.(2025){Wan Team}, Wang, Ai, Wen, Mao, Xie,
  et~al.]{wan}
{Wan Team}, A.~Wang, B.~Ai, B.~Wen, C.~Mao, C.-W. Xie, et~al.
\newblock {Wan: Open and Advanced Large-Scale Video Generative Models}.
\newblock \emph{arXiv preprint
  \href{https://arxiv.org/abs/2503.20314v1}{arXiv:2503.20314}}, 2025.

\bibitem[Wang et~al.(2023)Wang, Chen, Song, Ye, Liu, and Li]{genlvideo}
F.-Y. Wang, W.~Chen, G.~Song, H.-J. Ye, Y.~Liu, and H.~Li.
\newblock {Gen-L-Video: Multi-Text to Long Video Generation via Temporal
  Co-Denoising}.
\newblock \emph{arXiv preprint
  \href{https://arxiv.org/abs/2305.18264v1}{arXiv:2305.18264}}, 2023.

\bibitem[Wang et~al.(2025)Wang, Yuan, Zheng, Lin, Gao, Chen,
  et~al.]{spatialvid}
J.~Wang, Y.~Yuan, R.~Zheng, Y.~Lin, J.~Gao, L.-Z. Chen, et~al.
\newblock {SpatialVID: A Large-Scale Video Dataset with Spatial Annotations}.
\newblock \emph{arXiv preprint
  \href{https://arxiv.org/abs/2509.09676v1}{arXiv:2509.09676}}, 2025.

\bibitem[{WBench Team}(2026)]{wbenchboard}
{WBench Team}.
\newblock {WBench Navigation Leaderboard}, 2026.
\newblock URL \url{https://meituan-longcat.github.io/WBench/}.
\newblock Snapshot accessed September 9, 2026.

\bibitem[Wu et~al.(2026)Wu, He, Cheng, Yang, Zhang, Kang, Cai, Wei, Guo, Li,
  and Cheng]{infiniteworld}
R.~Wu, X.~He, M.~Cheng, T.~Yang, Y.~Zhang, Z.~Kang, X.~Cai, X.~Wei, C.~Guo,
  C.~Li, and M.-M. Cheng.
\newblock {Infinite-World: Scaling Interactive World Models to 1000-Frame
  Horizons via Pose-Free Hierarchical Memory}.
\newblock \emph{arXiv preprint
  \href{https://arxiv.org/abs/2602.02393v1}{arXiv:2602.02393}}, 2026.

\bibitem[Yang et~al.(2025)Yang, Huang, Chu, Xiao, Zhao, Wang, et~al.]{longlive}
S.~Yang, W.~Huang, R.~Chu, Y.~Xiao, Y.~Zhao, X.~Wang, et~al.
\newblock {LongLive: Real-time Interactive Long Video Generation}.
\newblock \emph{arXiv preprint
  \href{https://arxiv.org/abs/2509.22622v2}{arXiv:2509.22622}}, 2025.

\bibitem[Yin et~al.(2023)Yin, Gharbi, Zhang, Shechtman, Durand, Freeman, and
  Park]{dmd}
T.~Yin, M.~Gharbi, R.~Zhang, E.~Shechtman, F.~Durand, W.~T. Freeman, and
  T.~Park.
\newblock {One-step Diffusion with Distribution Matching Distillation}.
\newblock \emph{arXiv preprint
  \href{https://arxiv.org/abs/2311.18828v1}{arXiv:2311.18828}}, 2023.

\bibitem[Yin et~al.(2024{\natexlab{a}})Yin, Gharbi, Park, Zhang, Shechtman,
  Durand, and Freeman]{dmd2}
T.~Yin, M.~Gharbi, T.~Park, R.~Zhang, E.~Shechtman, F.~Durand, and W.~T.
  Freeman.
\newblock {Improved Distribution Matching Distillation for Fast Image
  Synthesis}.
\newblock \emph{arXiv preprint
  \href{https://arxiv.org/abs/2405.14867v1}{arXiv:2405.14867}},
  2024{\natexlab{a}}.

\bibitem[Yin et~al.(2024{\natexlab{b}})Yin, Zhang, Zhang, Freeman, Durand,
  Shechtman, and Huang]{causvid}
T.~Yin, Q.~Zhang, R.~Zhang, W.~T. Freeman, F.~Durand, E.~Shechtman, and
  X.~Huang.
\newblock {From Slow Bidirectional to Fast Autoregressive Video Diffusion
  Models}.
\newblock \emph{arXiv preprint
  \href{https://arxiv.org/abs/2412.07772v2}{arXiv:2412.07772}},
  2024{\natexlab{b}}.

\bibitem[Yin et~al.(2026)Yin, Wang, Zhan, Li, Zhang, and Zhao]{evoke}
Y.~Yin, G.~Wang, Y.~Zhan, C.~Li, K.~Zhang, and F.~Zhao.
\newblock {Alaya-EVOKE: From Linear-Scaling Supervision to Endless World}.
\newblock \emph{arXiv preprint
  \href{https://arxiv.org/abs/2608.13546v2}{arXiv:2608.13546}}, 2026.

\bibitem[Ying et~al.(2026)Ying, Hu, Ren, Li, Chen, Wang, et~al.]{wbench}
K.~Ying, H.~Hu, S.~Ren, J.~Li, F.~Chen, Z.~Wang, et~al.
\newblock {WBench: A Comprehensive Multi-turn Benchmark for Interactive Video
  World Model Evaluation}.
\newblock \emph{arXiv preprint
  \href{https://arxiv.org/abs/2605.25874v1}{arXiv:2605.25874}}, 2026.

\bibitem[Yuan et~al.(2026)Yuan, Yin, Li, Huang, Yang, and Yuan]{helios}
S.~Yuan, Y.~Yin, Z.~Li, X.~Huang, X.~Yang, and L.~Yuan.
\newblock {Helios: Real Real-Time Long Video Generation Model}.
\newblock \emph{arXiv preprint
  \href{https://arxiv.org/abs/2603.04379v1}{arXiv:2603.04379}}, 2026.

\bibitem[Zhao et~al.(2026)Zhao, Zhu, Zheng, Zhou, Yan, Li,
  et~al.]{causalforcingpp}
M.~Zhao, H.~Zhu, K.~Zheng, Z.~Zhou, B.~Yan, X.~Li, et~al.
\newblock {Causal Forcing++: Scalable Few-Step Autoregressive Diffusion
  Distillation for Real-Time Interactive Video Generation}.
\newblock \emph{arXiv preprint
  \href{https://arxiv.org/abs/2605.15141v1}{arXiv:2605.15141}}, 2026.

\bibitem[Zhu et~al.(2026)Zhu, Zhao, He, Su, Li, and Zhu]{causalforcing}
H.~Zhu, M.~Zhao, G.~He, H.~Su, C.~Li, and J.~Zhu.
\newblock {Causal Forcing: Autoregressive Diffusion Distillation Done Right for
  High-Quality Real-Time Interactive Video Generation}.
\newblock \emph{arXiv preprint
  \href{https://arxiv.org/abs/2602.02214v1}{arXiv:2602.02214}}, 2026.

\bibitem[Zhuang et~al.(2026)Zhuang, Zhang, Bian, Li, Luo, Liu, et~al.]{sgf}
J.~Zhuang, S.~Zhang, Y.~Bian, Y.~Li, Y.~Luo, Y.~Liu, et~al.
\newblock {Self Gradient Forcing: Native Long Video Extrapolation}.
\newblock \emph{arXiv preprint
  \href{https://arxiv.org/abs/2607.20368v1}{arXiv:2607.20368}}, 2026.

\end{thebibliography}
\endgroup

\end{document}